\documentclass[sigconf]{acmart}

\usepackage{amsmath,amsfonts,bm}

\def\eqref#1{equation~\ref{#1}}

\def\1{\bm{1}}

\DeclareMathAlphabet{\mathsfit}{\encodingdefault}{\sfdefault}{m}{sl}
\SetMathAlphabet{\mathsfit}{bold}{\encodingdefault}{\sfdefault}{bx}{n}

\usepackage[font=small,skip=0pt]{caption}
\usepackage{graphicx,adjustbox}
\usepackage[linesnumbered,ruled,vlined]{algorithm2e}
\usepackage{multirow}
\usepackage{diagbox}
\usepackage{booktabs}
\usepackage{enumitem}
\usepackage[normalem]{ulem}
\useunder{\uline}{\ul}{}
\SetKwComment{Comment}{/* }{ */}
\SetAlFnt{\small}
\SetKwInput{kwInput}{Input}
\SetKwInput{kwOutput}{Output}
\SetKwComment{LineComment}{\# }{\#}

\AtBeginDocument{%
  }

\copyrightyear{2026}
\acmYear{2026}
\setcopyright{cc}
\setcctype{by}
\acmConference[KDD '26]{Proceedings of the 32nd ACM SIGKDD Conference on Knowledge Discovery and Data Mining V.2}{August 09--13, 2026}{Jeju Island, Republic of Korea}
\acmBooktitle{Proceedings of the 32nd ACM SIGKDD Conference on Knowledge Discovery and Data Mining V.2 (KDD '26), August 09--13, 2026, Jeju Island, Republic of Korea}
\acmDOI{10.1145/3770855.3818201}
\acmISBN{979-8-4007-2259-2/2026/08}
\renewcommand{\shortauthors}{Xuan-May Le et al.}

\begin{document}

    \title{Efficient Test-Time Scaling for LLM-based Time Series Forecasting}


\author{Xuan-May Le}
\authornote{Both authors contributed equally to this research.}
\email{xuanmay.le@student.unimelb.edu.au}
\affiliation{%
  \institution{The University of Melbourne}
  \city{Melbourne}
  \state{Victoria}
  \country{Australia}
}

\author{Minh-Tuan Tran}
\authornotemark[1]
\email{tuan.tran7@monash.edu}
\affiliation{%
  \institution{Monash University}
  \city{Melbourne}
  \state{Victoria}
  \country{Australia}
}

\author{Ling Luo}
\email{ling.luo@unimelb.edu.au}
\affiliation{%
  \institution{The University of Melbourne}
  \city{Melbourne}
  \state{Victoria}
  \country{Australia}
}

\author{Uwe Aickelin}
\email{uwe.aickelin@unimelb.edu.au}
\affiliation{%
  \institution{The University of Melbourne}
  \city{Melbourne}
  \state{Victoria}
  \country{Australia}
}

\author{Dinh Phung}
\email{dinh.phung@monash.edu}
\affiliation{%
  \institution{Monash University}
  \city{Melbourne}
  \state{Victoria}
  \country{Australia}
}

\author{Trung Le}
\email{trunglm@monash.edu}
\affiliation{%
  \institution{Monash University}
  \city{Melbourne}
  \state{Victoria}
  \country{Australia}
}








\renewcommand{\shortauthors}{Xuan-May Le et al.}

\begin{abstract}

Long-term time series forecasting benefits from preserving global structure such as trends and seasonality. Recent LLM-based forecasters often improve accuracy through test-time scaling (e.g., iterative refinement), but these methods are computationally expensive and increasingly prone to global-shape mismatch as the prediction horizon extends. We propose SCALER, a coarse-to-fine forecasting framework that first employs a lightweight Transformer tailored to long-term shape modeling to predict a coarse representation of future dynamics. This predicted shape then serves as a compact guide for an LLM to perform test-time scaling via iterative coarse-to-fine residual token refinement, while processing substantially fewer tokens at each step. By guiding refinement with an explicit future-shape prediction, SCALER reduces reliance on long description prompts, and its fixed-step refinement avoids costly reward-model-based selection, further lowering computational overhead. Experimental results demonstrate that SCALER outperforms strong forecasting baselines in long-term, short-term and zero-shot forecasting while significantly reducing the inference cost associated with scaled LLM for time series forecasting. Code: \url{https://github.com/xuanmay2701/SCALER}.

\end{abstract}




\maketitle

\section{Introduction}

\begin{figure}[t]
\centering
\begin{center}
\includegraphics[width=\linewidth]{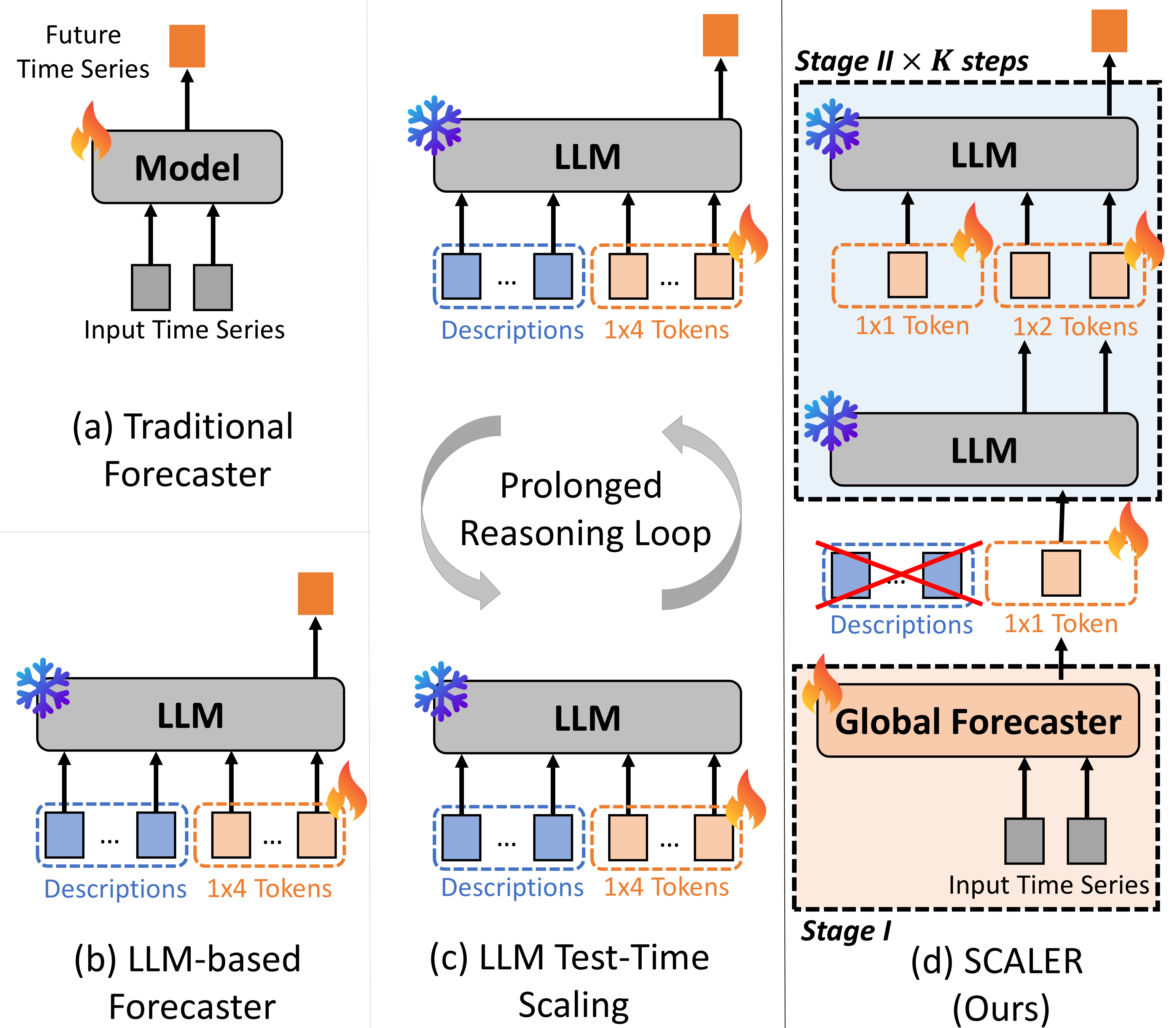}
\end{center}
\caption{(a) Standard: direct forecasting; may lose long-horizon trend/seasonality.
(b) LLM forecaster: prompt-based prediction; long horizons need long prompts and can drift in shape.
(c) LLM + test-time scaling: long iterative refinement improves accuracy but is costly and can amplify shape mismatch.
(d) SCALER (ours): predicts coarse future shape first, then undertakes fixed-step residual refinement with fewer tokens for cheap, shape-consistent forecasts.}
\label{fig:moti}
\end{figure}

Time series data consists of sequential observations indexed over time \cite{arima,arma,ship,pisd,medformer,shapeformer,ppsn}, and forecasting such sequences plays a central role in many real-world domains, such as finance \cite{finance-1,finance-2}, healthcare \cite{heathcare-1,heathcare-2}, weather prediction \cite{weather-forecast-1,weather-forecast-2}, and energy demand modeling \cite{energy-1,energy-2}, where reliable future estimates directly support decision-making and resource allocation. However, many conventional forecasting approaches face difficulties capturing complex temporal dynamics at multiple scales, often failing to jointly represent long-range dependencies and fine-grained local variations in the data.


Over the past decades, numerous approaches have been proposed for time series forecasting. Traditional deep-learning-based models \cite{timexer,timemixer,patchtst,itransformer,timesnet} (Figure~\ref{fig:moti}a) are effective at capturing global temporal patterns such as trends and seasonality through nonlinear architectures, and achieve strong performance on standard benchmarks. More recently, large language model (LLM) forecasters \cite{llm-tsf1,llm-tsf2,llm-timecma,llm-lvicl,time-llm} (Figure~\ref{fig:moti}b) have attracted renewed interest, bringing multimodal understanding and generative modeling perspectives to time series forecasting. Despite these advances, most LLM-based TSF methods still formulate forecasting as direct sequence-to-sequence mapping, producing predictions in a single inference pass without explicit intermediate reasoning. Consequently, scaling-time inference strategies \cite{timereasoner} (Figure~\ref{fig:moti}c) have been explored to improve accuracy by generating multiple candidates or iteratively refining predictions, but they face two key challenges: (1) iterative refinement and candidate selection can be computationally expensive, causing inference cost to grow rapidly with horizon length and desirable accuracy; and (2) repeated local corrections can accumulate into a global-shape drift, distorting trends and seasonality as the horizon extends. These issues are often exacerbated by reliance on long prompts or expensive reward-model-based candidate selection.

To address the above issues, we introduce SCALER, a two-stage coarse-to-fine framework (Figure~\ref{fig:moti}d) designed for efficient test-time scaling in long-horizon forecasting. SCALER first tokenises the history into a compact multi-scale representation using a shared encoder, which amortises feature extraction and prevents repeated processing of raw inputs. In Stage I, a lightweight forecaster predicts an explicit coarse future shape that captures the horizon’s low-frequency structure (trend, seasonality, and regime changes). In Stage II, we allocate additional test-time computation through a fixed K-step refinement procedure driven by a pretrained LLM. The LLM’s pretrained generative prior improves robustness and coherence when refining under limited supervision. Conditioned on the history tokens and the coarse-shape anchor, it generates finer-scale token blocks that progressively enrich local details without deviating from the intended global trajectory. The final forecast is produced by decoding the refined token context back to the original resolution. This design scales inference with a fixed, deployment-friendly budget, avoids reward-model-based candidate selection, and mitigates global-shape mismatch that becomes more severe as the horizon grows.


This design provides two practical advantages. \textbf{(i) Efficiency:} SCALER offloads global-structure prediction to a lightweight forecaster and uses the LLM only for incremental refinement over compact token blocks, substantially reducing the token-processing burden of scaled inference. In avarge inference time, our SCALER is 7 times faster than standard test-time scaling forecasters. \textbf{(ii) Stability:} anchoring refinement to an explicit coarse shape constrains the LLM’s updates, reducing reliance on long prompts and limiting long-horizon drift. Moreover, the fixed-step refinement schedule yields deterministic test-time computation and avoids costly reward-model-based candidate selection, making inference predictable and deployment-friendly.

Our main contributions are:
\begin{itemize}[leftmargin=*, nosep]
    \item We propose \textsc{SCALER}, a two-stage framework for \emph{test-time scaling} in long-horizon forecasting: a lightweight forecaster predicts a coarse future shape, and a pretrained LLM refines it into full-resolution forecasts.
    \item We introduce multi-scale tokenization with a \emph{fixed-step} coarse-to-fine refinement schedule, where each scaling step generates only a compact token block conditioned on the shape anchor, reducing computational cost and limiting long-horizon drift.
    \item Experiments across long-term, short-term, and zero-shot benchmarks show that \textsc{SCALER} improves accuracy over strong baselines while cutting the inference cost compared to standard test-time scaling LLM-based time series forecasters.

\end{itemize}

\section{Related Work}

\noindent\textbf{Time series forecasting.}
A wide range of TSF methods span statistical, deep learning, and generative paradigms. Classical models such as ARMA \cite{arma} and ARIMA \cite{arima} mainly capture linear dependencies, while Transformer variants including Informer \cite{informer}, Autoformer \cite{autoformer}, Crossformer \cite{crossformer}, and Pyraformer \cite{pyraformer} use attention to model long-range temporal structure. PatchTST \cite{patchtst} improves efficiency by patchifying sequences, and iTransformer \cite{itransformer} captures multivariate interactions via transposed attention. Lightweight linear baselines (e.g., DLinear \cite{dlinear}, RLinear \cite{rlinear}, TimeMixer \cite{timemixer}, TimeXer \cite{timexer}) remain competitive on large benchmarks. Overall, these deterministic forecasters are typically effective at preserving coarse global dynamics (trends/seasonality), which becomes increasingly important for long horizons.

Generative TSF is another active direction. Diffusion-style methods such as TimeGrad \cite{timegrad}, D3VAE \cite{d3vae}, and TSDiff \cite{tsdiff} synthesise futures via iterative generation, and ARMD \cite{armd} predicts future chunks with an autoregressive moving-window strategy. While such models can better recover local variability and uncertainty, iterative generation can accumulate errors and drift in global structure over long horizons \cite{timegrad,d3vae,tsdiff}, motivating approaches that explicitly anchor long-term shape while refining details.

\noindent\textbf{LLM-based and foundation-model forecasting.}
LLM-based TSF has grown rapidly, bringing multimodal understanding and generative modeling perspectives into forecasting. TimeLLM \cite{time-llm} reprograms time-series inputs to match a pre-trained LLM interface, PromptCast \cite{llm-promptcast} uses prompt-driven textualisation, LLMTime \cite{llmtime} studies numeric token encodings, and AutoTimes \cite{llm-autotimes} aligns forecasting with next-token prediction. Other methods adapt parameters for better temporal alignment, including fine-tuning (FPT \cite{llm-fpt}, LLM4TS \cite{llm4ts}) and lightweight updates (CALF \cite{llm-calf}), while LVICL \cite{llm-lvicl} strengthens ICL by injecting learned vectors. However, most LLM-based approaches still treat forecasting as direct sequence mapping (often in a single pass) and rely on long prompts or adaptation pipelines that increase compute/memory. Our method instead keeps LLM parameters frozen and targets efficient, shape-faithful long-horizon forecasting.

\noindent\textbf{Test-time scaling and iterative refinement.}
Recent “slow-thinking” inference improves TSF by scaling test-time compute, e.g., sampling multiple futures, selecting candidates, or iteratively refining predictions \cite{timereasoner}. While effective, these strategies face two key issues for TSF: (1) cost grows quickly with horizon length because each round reprocesses long sequences and may require expensive selection (often via reward models); and (2) repeated local edits can compound into global-shape drift, distorting trends and seasonal phase as horizons extend. These limitations motivate our decoupled design that anchors refinement to an explicit long-horizon structure while enabling efficient fixed-step scaling.


\section{Proposed Method}
\label{sec:method}

\begin{figure*}[t]
\centering
\begin{center}
\includegraphics[width=\linewidth]{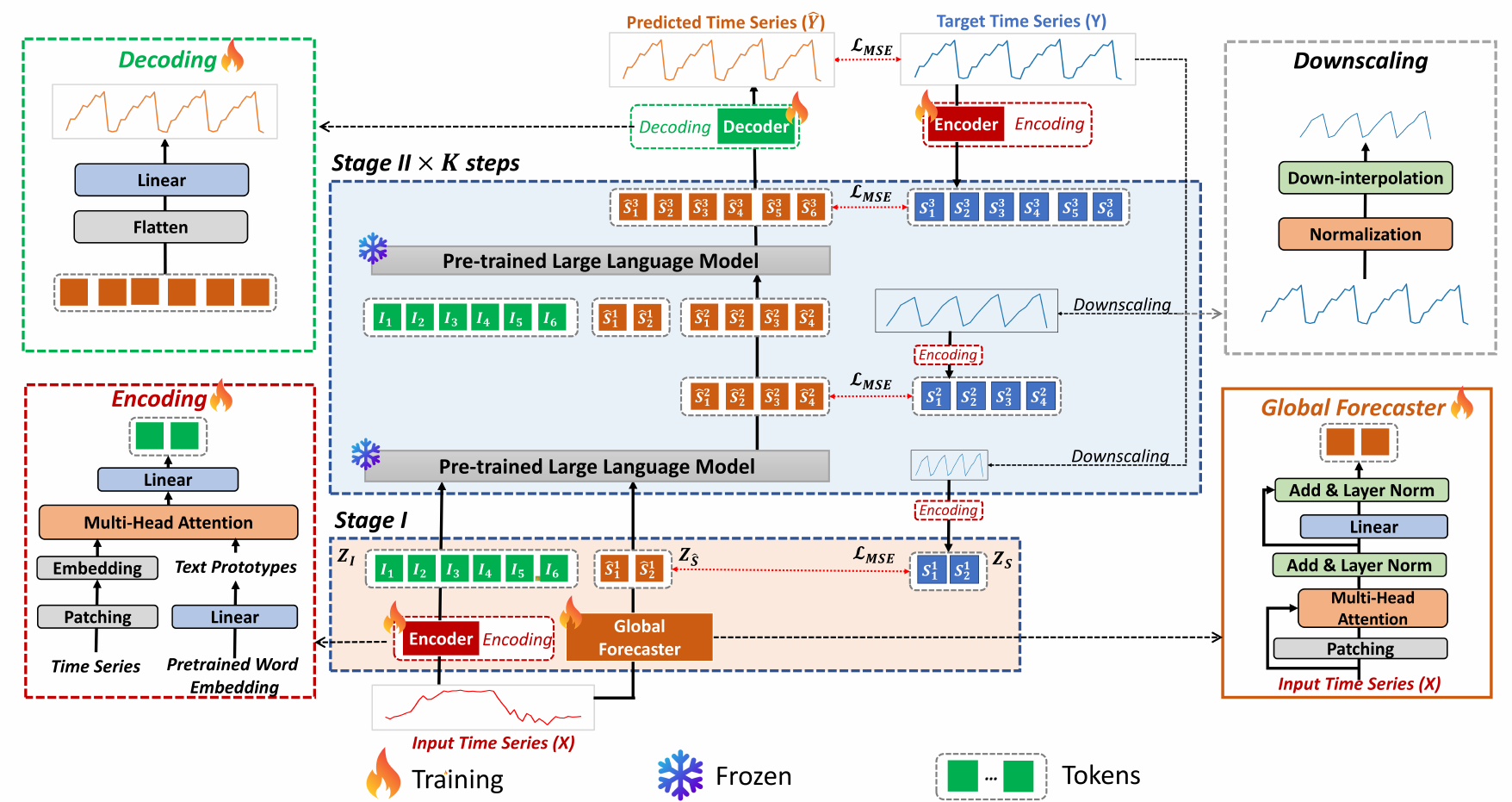}
\end{center}
\caption{\textbf{SCALER architecture.} Given the input/history time series $X$, a lightweight coarse forecaster $\mathcal{F}$ predicts a downscaled future shape token $\hat{S}^1$ that captures low-frequency dynamics. A shared multi-scale encoder $\mathcal{E}$ maps $X$ into a continuous token sequence $Z_I$. Conditioned on $Z_I$ and a growing set of shape tokens initialised as $Z_S = [S^1]$, an LLM refiner $\mathcal{G}$ runs a fixed $K$-step coarse-to-fine loop: at step $k$, it predicts a finer-scale token block $S^k$ and appends it to the context $Z_S$ for the next step. The final forecast is obtained by decoding the finest-scale tokens, $S=\mathcal{D}\!\left(Z_S^{(K)}\right)$, yielding shape-faithful refinement with predictable inference cost and no candidate selection.
}

\label{fig:arch}
\end{figure*}

\subsection{Problem Setup}
We study multivariate forecasting. Given a historical sequence with $ C $ variables, 
$X=\{x_t\}_{t=1}^{L}$ and $x_t\in\mathbb{R}^{C}$, our goal is to predict the future horizon
$Y=\{y_t\}_{t=L+1}^{L+T}$. In that, $L$ and $T$ are the lengths of history and future time series, respectively. We seek forecasts that preserve \emph{global structure} (e.g., trends, seasonality, regime shifts) while recovering \emph{local details} under a limited test-time budget.

\subsection{Overview of {SCALER}}
\label{sec:overview}
SCALER is a coarse-to-fine framework with two stages (refer to Figure \ref{fig:arch}).

\noindent\textbf{Stage I: Coarse-shape prediction.}
A lightweight forecaster $\mathcal{F}$ predicts a downscaled future shape token $\hat{S}^{1}$ that
summarises low-frequency dynamics over the full horizon.

\noindent\textbf{Stage II: Fixed-step multi-scale refinement.}
We encode the history into tokens $Z_I=\mathcal{E}(X)$ and initialise the conditioning context as
$Z_S=[\hat{S}^{1}]$. A refiner $\mathcal{G}$ then performs a fixed $K$-step residual refinement from coarse to fine:
at step $k$, it predicts a finer-scale token block $S^{k}$ conditioned on $(Z_I, Z_S)$, and appends it to
$Z_S$ for the next step. This yields a shape-faithful refinement procedure with predictable inference cost
and no candidate selection.

\subsection{Multi-Scale Patch Encoding}
\label{sec:ms_encoder}

We encode each time series into a sequence of \emph{embedding tokens}.

\noindent\textbf{Encoding.} Given a multivariate series $X\in\mathbb{R}^{L\times C}$, we patchify the input and encode it into a history token sequence $Z_I=[I_1,\ldots,I_m]$:
\begin{equation}
Z_I = [\,I_1, \ldots, I_m\,] = \text{Encoding}(X) = \mathcal{E}\!\left(\mathrm{Patching}(X;\ell)\right),
\end{equation}
where $\ell$ is the patch length, $m$ is number of tokens, $\mathcal{E}$ is a patch encoder that aligns time-series patches with the LLM embedding space via prototype-based cross-attention. We construct a compact prototype set by linearly projecting the LLM word-embedding matrix $W$,
\begin{equation}
P=\mathrm{Linear}(W),
\end{equation}
and encode each patch by cross-attending its query (from patch token $I$) to prototypes $P$ as keys and values:
\begin{equation}
Z=\mathrm{Softmax}\!\left(\frac{QK^{\top}}{\sqrt{d}}\right)V,
\end{equation}
where $Q$ is projected from $I$, and $K,V$ are projected from $P$.

\noindent\textbf{Downscaling.}
Let $\{\mathrm{Scale}_k\}_{k=1}^{K}$ be ordered from coarse (e.g. 12 and 24 timestamps) to fine (e.g. 48 and 96 timestamps). For the future horizon $Y\in\mathbb{R}^{T\times C}$, we optionally construct a downscaled view at scale $k$ via a normalise-then-interpolate operator:
\begin{align}
Y^{k} &= \text{Downscaling}(Y;\mathrm{Scale}_k) \\
&= \text{Down-interpolating}\!\left(\mathrm{Norm}(Y),\, \mathrm{Scale}_k\right),
\end{align}
where $\mathrm{Norm}(\cdot)$ normalises each channel (e.g., by mean and std), and $\text{Down-interpolating}(\cdot)$ downscales the sequence by interpolating to the target resolution (e.g., linear interpolation). We then obtain the token sequence by the same patch-encoding operator:
\begin{equation}
Z_S = [\,S^{k}_1, S^{k}_2, \ldots, S^{k}_{n_k}\,]
= \text{Encoding}(Y^{k}),
\qquad S^{k}_j \in \mathbb{R}^{D},
\end{equation}
where $n_k$ is the number of patches (i.e., tokens) obtained after patching $S^{k}$.

For multivariate inputs we apply this per channel and concatenate token streams across channels with lightweight identifiers to form the final conditioning sequence.

\subsection{Stage I: Global Coarse-Shape Forecaster}
\label{sec:shape}

Given the history $X=x_{1:L}$, we first predict an explicit \emph{coarse global future shape} that captures low-frequency dynamics over the horizon. A compact transformer model $\mathcal{F}$ predicts the coarse shape from the history:
\begin{equation}
\hat{S}^{1} = \mathcal{F}(X),
\end{equation}
where $\mathcal{F}$ is intentionally lightweight since it only models global structure
(trends/seasonality/regime) at a reduced resolution.

We next construct the coarse target by applying our normalise-then-interpolate downscaling operator to the future segment $Y=I_{L+1:L+T}$:
\begin{align}
S^{1} &= \mathrm{Downscaling}(Y;\mathrm{Scale}_1)
\end{align}
We then obtain its coarse-shape token sequence by joint patch encoding:
\begin{equation}
Z_{S^1} =[\,S^{1}_{1}, S^{1}_{2}, \ldots, S^{1}_{n_1}\,] = \text{Encoding}(S^1)
\end{equation}

\noindent\textbf{Shape supervision.}
We train $\mathcal{F}$ with a regression loss against the coarse target:
\begin{equation}
\mathcal{L}_{\text{shape}} = \mathcal{L}_{\text{MSE}}(\hat{S}^{1}, S^1) = \lVert \hat{S}^{1} - S^{1} \rVert_{2}^{2},
\end{equation}
(with $\mathcal{L}_\text{MSE}$ as a drop-in alternative). The predicted coarse shape $\hat{S}^{1}$ initialises the
conditioning context for the subsequent fixed-step coarse-to-fine refinement.

\subsection{Stage II: Fixed-Step Multi-Scale Refinement}
\label{sec:refine}

The predicted coarse shape $\hat{S}^{1}$ captures global structure but omits fine-grained details.
We implement \emph{test-time scaling} through a fixed $K$-step coarse-to-fine refinement loop, where
additional inference compute is allocated to progressively finer temporal scales.
Crucially, using a pretrained LLM $\mathcal{G}$ as the refiner injects a strong sequence prior,
which helps produce coherent long-horizon completions and stabilizes refinement when supervision is limited.

\noindent\textbf{Token initialisation.}
We encode the history once to obtain the history token sequence:
\begin{equation}
Z_I = [\,I_1, I_2, \ldots, I_m\,] = \text{Encoding}(X).
\end{equation}
We initialise the refinement context with the coarse-shape tokens predicted in Stage~I:
\begin{equation}
Z_S = [\,S^{1}\,], \qquad \text{where } S^{1}=\text{Encoding}(\hat{S}^{1})).
\end{equation}

\noindent\textbf{Test-time scaling policy (multi-scale refinement).}
We treat $\{\mathrm{Scale}_k\}_{k=1}^{K}$ as a test-time scaling schedule that determines how refinement process is distributed across resolutions. During training, we form the scale-$k$ target by a normalise-then-interpolate downscaling:
\begin{align}
S^{k} &= \mathrm{Downscaling}(Y;\mathrm{Scale}_k) 
\end{align}
We then obtain its token sequence:
\begin{equation}
Z_{S^k}=[\,S^{k}_1,\ldots,S^{k}_{n_k}\,] = \text{Encoding}(Y^{k}).
\end{equation}
At step $k$, the refiner pretrained large language model $\mathcal{G}$ predicts the scale-$k$ token block conditioned on the history tokens and the coarser context previously predicted:
\begin{equation}
\hat{S}^{k} = \mathcal{G}\!\left(Z_I,\, Z_S\right),
\end{equation}
and we append it to the context for the next step:
\begin{equation}
Z_S \;\leftarrow\; [\,Z_S,\; \hat{S}^{k}\,].
\end{equation}

\begin{algorithm}[t]
\caption{{SCALER} Training}
\label{alg:scaler_train}
\KwIn{History $X$, future $Y$, patch length $\ell$, scales $\{\mathrm{Scale}_1,\ldots,\mathrm{Scale}_K\}$, refinement steps $K$, weights $\lambda,\beta$}
\KwOut{Predicted time series $\hat{Y}$}
\BlankLine
\tcp{{\color{blue}Stage I: coarse shape forecasting}}
$S^{1} \leftarrow \text{Downscaling}\!\left(Y,\,\mathrm{Scale}_1\right)$\;
$\hat{S}^{1} \leftarrow \mathcal{F}(X)$ \;
$\mathcal{L}_{\text{shape}} \leftarrow \lVert \hat{S}^{1} - S^{1} \rVert_{2}^{2}$ \;
\BlankLine

\tcp{{\color{blue}Stage II: 
iterative refining}}
$Z_I \leftarrow \text{Encoding}(X)$ \;
$Z_S \leftarrow \big[\text{Encoding}(\hat{S}^{1})\big]$ \;
$\mathcal{L}_{\text{refine}} \leftarrow 0$ \;
\For{$k=1$ {\bf to} $K$}{ 
    $Y^{k} \leftarrow \text{Downscaling}\!\left(Y,\,\mathrm{Scale}_k\right)$ 
    $\hat{S}^{k} \leftarrow \mathcal{G}(Z_I, Z_S)$ \tcp*[r]{predict scale-$k$ tokens}
    $\mathcal{L}_{\text{refine}} \leftarrow \mathcal{L}_{\text{refine}}
    + \left\lVert \hat{S}^{k} - \text{Encoding}(Y^{k}) \right\rVert_{2}^{2}$ \;
    $Z_S \leftarrow [\,Z_S,\; \hat{S}^{k}\,]$ \;
}
$\hat{Y} \leftarrow \text{Decoding}(Z_S)$ \;
$\mathcal{L}_{\text{ts}} \leftarrow \lVert \hat{Y} - Y \rVert_{2}^{2}$ \;
\BlankLine

$\mathcal{L} \leftarrow \mathcal{L}_{\text{shape}} + \lambda\,\mathcal{L}_{\text{refine}} + \beta\,\mathcal{L}_{\text{ts}}$ \;
\Return{$\hat{Y}$}\;
\end{algorithm}

\begin{algorithm}[t]
\caption{{SCALER} Inference}
\label{alg:scaler_infer}
\KwIn{History $X$, patch length $\ell$, refinement steps $K$}
\KwOut{Forecast $\hat{Y}$}
\BlankLine
\tcp{\color{blue}Stage I: coarse shape forecasting}
$\hat{S}^{1} \leftarrow \mathcal{F}(X)$ \; 

\tcp{\color{blue}Stage II: iterative refining}
$Z_I \leftarrow \text{Encoding}(X)$ \tcp*[r]{history tokens}
$Z_S \leftarrow \Big[\,\text{Encoding}(\hat{S}\Big)]$ \tcp*[r]{init context}
\For{$k=1$ {\bf to} $K$}{
    $\hat{S}^{k} \leftarrow \mathcal{G}(Z_I, Z_S)$ \tcp*[r]{predict finer-scale tokens}
    $Z_S \leftarrow [\,Z_S,\; \hat{S}^{k}\,]$ \tcp*[r]{append for next step}
}
$\hat{Y} \leftarrow \text{Decoding}(Z_S)$ \tcp*[r]{decode final tokens}
\Return{$\hat{Y}$}\;
\end{algorithm}

\begin{table*}[t]

\centering
\caption{\textbf{Long-term forecasting results (averaged over horizons).} We report MSE/MAE on seven benchmarks. \textbf{Bold} denotes the best result and underline denotes the second best within each dataset. The last row reports the number of first-place results across all dataset--horizon pairs. All results are taken from \cite{llm-lvicl}, and we rerun TimeReasoner under our setup. All numbers are averaged over five runs with different random seeds.}
\label{tab:scaler_full_ltsf}
\setlength{\tabcolsep}{2.2pt}
\renewcommand{\arraystretch}{1.05}
\begin{adjustbox}{width=\textwidth}
\begin{tabular}{@{}l|cc|cc|cc|cc|cc|cc|cc|cc|cc|cc|cc|cc@{}}
\toprule
\multicolumn{1}{c}{}        & \multicolumn{4}{|c}{Test-Time Scaling}                                                                 & \multicolumn{12}{|c}{LLM-based Model}                                                                                                                                                                                                                                                                             & \multicolumn{8}{|c}{Other Deep Learning Model}                                                                                                                                                               \\ \midrule
\multicolumn{1}{c}{}        & \multicolumn{2}{|c}{SCALER (Ours)}                 & \multicolumn{2}{c}{TimeReasoner}                  & \multicolumn{2}{|c}{LVICL}                         & \multicolumn{2}{c}{AutoTimes}                     & \multicolumn{2}{c}{TimeLLM}                       & \multicolumn{2}{c}{FPT}                           & \multicolumn{2}{c}{Chronos}                       & \multicolumn{2}{c}{TimeFM}                        & \multicolumn{2}{|c}{iTransformer}                  & \multicolumn{2}{c}{DLinear}                       & \multicolumn{2}{c}{PatchTST}                      & \multicolumn{2}{c}{TimesNet}                      \\ \midrule
\multicolumn{1}{l}{Dataset} & \multicolumn{1}{|c}{MSE} & \multicolumn{1}{c}{MAE} & \multicolumn{1}{|c}{MSE} & \multicolumn{1}{c}{MAE} & \multicolumn{1}{|c}{MSE} & \multicolumn{1}{c}{MAE} & \multicolumn{1}{|c}{MSE} & \multicolumn{1}{c}{MAE} & \multicolumn{1}{|c}{MSE} & \multicolumn{1}{c}{MAE} & \multicolumn{1}{|c}{MSE} & \multicolumn{1}{c}{MAE} & \multicolumn{1}{|c}{MSE} & \multicolumn{1}{c}{MAE} & \multicolumn{1}{|c}{MSE} & \multicolumn{1}{c}{MAE} & \multicolumn{1}{|c}{MSE} & \multicolumn{1}{c}{MAE} & \multicolumn{1}{|c}{MSE} & \multicolumn{1}{c}{MAE} & \multicolumn{1}{|c}{MSE} & \multicolumn{1}{c}{MAE} & \multicolumn{1}{|c}{MSE} & \multicolumn{1}{c}{MAE} \\ \midrule
ETTh1                       & \textbf{0.376}          & \textbf{0.408}          & 0.403                   & 0.417                   & {\ul 0.381}             & {\ul 0.412}             & 0.389                   & 0.422                   & 0.408                   & 0.423                   & 0.427                   & 0.426                   & 0.401                   & 0.434                   & 0.417                   & 0.451                   & 0.438                   & 0.45                    & 0.423                   & 0.437                   & 0.413                   & 0.431                   & 0.458                   & 0.45                    \\
ETTh2                       & \textbf{0.321}          & \textbf{0.373}          & 0.348                   & 0.389                   & {\ul 0.326}             & {\ul 0.376}             & 0.352                   & 0.395                   & 0.354                   & 0.393                   & 0.353                   & 0.391                   & 0.327                   & 0.377                   & 0.334                   & 0.385                   & 0.382                   & 0.414                   & 0.431                   & 0.446                   & 0.33                    & 0.379                   & 0.414                   & 0.427                   \\
ETTm1                       & {\ul 0.324}             & 0.374                   & 0.344                   & {\ul 0.373}             & 0.328                   & 0.378                   & 0.332                   & 0.38                    & 0.35                    & 0.378                   & 0.366                   & 0.382                   & \textbf{0.322}          & \textbf{0.371}          & 0.341                   & 0.393                   & 0.37                    & 0.399                   & 0.357                   & 0.378                   & 0.351                   & 0.38                    & 0.4                     & 0.406                   \\
ETTm2                       & \textbf{0.234}          & \textbf{0.302}          & 0.245                   & 0.311                   & {\ul 0.239}             & {\ul 0.306}             & 0.243                   & 0.31                    & 0.254                   & 0.314                   & 0.265                   & 0.315                   & 0.245                   & 0.313                   & 0.242                   & 0.309                   & 0.272                   & 0.331                   & 0.267                   & 0.333                   & 0.255                   & 0.315                   & 0.291                   & 0.333                   \\
ECL                         & {\ul \textbf{0.154}}    & \textbf{0.245}          & {\ul \textbf{0.154}}    & 0.25                    & 0.158                   & {\ul 0.248}             & 0.159                   & 0.253                   & 0.159                   & 0.253                   & 0.167                   & 0.263                   & 0.164                   & 0.258                   & 0.162                   & 0.255                   & 0.161                   & 0.256                   & 0.177                   & 0.274                   & 0.159                   & 0.253                   & 0.192                   & 0.295                   \\
Weather                     & \textbf{0.217}          & {\ul 0.257}             & {\ul 0.218}             & \textbf{0.254}          & 0.224                   & 0.262                   & 0.235                   & 0.273                   & 0.225                   & {\ul 0.257}             & 0.237                   & 0.27                    & 0.22                    & 0.263                   & 0.229                   & 0.27                    & 0.238                   & 0.272                   & 0.24                    & 0.3                     & 0.226                   & 0.264                   & 0.259                   & 0.287                   \\
Traffic                     & \textbf{0.367}          & \textbf{0.256}          & 0.382                   & {\ul 0.259}             & {\ul 0.37}              & 0.26                    & 0.374                   & 0.264                   & 0.388                   & 0.264                   & 0.414                   & 0.294                   & 0.376                   & 0.264                   & 0.383                   & 0.269                   & 0.379                   & 0.272                   & 0.434                   & 0.295                   & 0.391                   & 0.264                   & 0.62                    & 0.336                   \\ \midrule
1st Count                   & \textbf{6}              & \textbf{5}              & {\ul 1}                 & {\ul 1}                 & 0                       & 0                       & 0                       & 0                       & 0                       & 0                       & 0                       & 0                       & {\ul 1}                 & {\ul 1}                 & 0                       & 0                       & 0                       & 0                       & 0                       & 0                       & 0                       & 0                       & 0                       & 0                       \\ \bottomrule
\end{tabular}
\end{adjustbox}

\end{table*}

We supervise refinement at two levels. First, we define $\mathcal{L}_{\text{refine}}$ as a scale-wise $\mathcal{L}_\text{MSE}$ loss on the downscaled targets at level $k$:
\begin{equation}
\mathcal{L}_{\text{refine}} = \sum_{k=1}^{K} \left\lVert \hat{S}^{k} - S^{k} \right\rVert_{2}^{2}.
\end{equation}
\noindent\textbf{Decoding.} Second, we decode the final token set into a full-resolution forecast using a decoder $\mathcal{D}$:
\begin{equation}
\hat{Y} = \text{Decoding}(Z_S) = \mathcal{D}(Z_S),
\end{equation}
and impose a $\mathcal{L}_\text{MSE}$ loss in the original time-series space:
\begin{equation}
\mathcal{L}_{\text{ts}} = \mathcal{L}_{\text{MSE}}(\hat{Y}, Y) = \left\lVert \hat{Y} - Y \right\rVert_{2}^{2}.
\end{equation}
The overall training objective is
\begin{equation}
\mathcal{L} = \mathcal{L}_{\text{shape}} \;+\; \lambda\,\mathcal{L}_{\text{refine}} \;+\; \beta\,\mathcal{L}_{\text{ts}},
\end{equation}
where $\lambda$ and $\beta$ balance token-level refinement and time-series reconstruction.

\noindent\textbf{Inference.}
At test time, we first predict $\hat{S}^{1}=\mathcal{F}(X)$, initialise $Z_S=[S^{1}]$, and run the fixed
$K$-step loop to obtain the final token set $Z_S$. We then decode $\hat{Y}=\mathcal{D}(Z_S)$ (optionally
upsampling if needed), yielding predictable inference cost without candidate generation or selection. Algorithm~\ref{alg:scaler_train} gives the training
procedure, while Algorithm~\ref{alg:scaler_infer} details the inference objective.

\section{Experiments}

\begin{table*}[t]

\centering
\caption{\textbf{Short-term forecasting results.} We report sMAPE, MASE, and OWA on four scenario groups (Year, Quarter, Month, Others) and their average, where lower is better. \textbf{Bold} denotes the best result and underline denotes the second best within each row. The last row reports the number of first-place results across all scenario--metric combinations. All numbers are averaged over five runs with different random seeds.}
\label{tab:stf_results}
\setlength{\tabcolsep}{3.2pt}
\renewcommand{\arraystretch}{1.05}
\begin{adjustbox}{width=\textwidth}
\begin{tabular}{@{}l|l|ccccccccc@{}}
\toprule
Scenario & Category  & SCALER (ours)        & TimeReasoner & LVICL       & AutoTimes & TimeLLM & FPT    & DLinear & PatchTST & TimesNet \\ \midrule
         & sMAPE     & \textbf{12.984}      & {\ul 13.059} & 13.101      & 13.319    & 13.419  & 13.531 & 13.866  & 13.517   & 13.394   \\
Year     & MASE      & \textbf{2.977}       & {\ul 2.983}  & 2.988       & 2.993     & 3.005   & 3.015  & 3.006   & 3.031    & 3.004    \\
         & OWA       & \textbf{0.764}       & {\ul 0.771}  & 0.775       & 0.784     & 0.789   & 0.793  & 0.802   & 0.795    & 0.787    \\ \midrule
         & sMAPE     & \textbf{9.82}        & {\ul 9.935}  & 10.01       & 10.101    & 10.11   & 10.177 & 10.689  & 10.847   & 10.101   \\
Quarter  & MASE      & \textbf{1.159}       & {\ul 1.163}  & 1.17        & 1.182     & 1.178   & 1.194  & 1.294   & 1.315    & 1.183    \\
         & OWA       & \textbf{0.872}       & {\ul 0.882}  & {\ul 0.882} & 0.89      & 0.889   & 0.897  & 0.957   & 0.972    & 0.89     \\ \midrule
         & sMAPE     & \textbf{12.195}      & {\ul 12.311} & 12.321      & 12.71     & 12.98   & 12.894 & 13.372  & 14.584   & 12.866   \\
Month    & MASE      & \textbf{0.916}       & {\ul 0.93}   & 0.934       & 0.934     & 0.963   & 0.956  & 1.014   & 1.169    & 0.964    \\
         & OWA       & \textbf{0.856}       & {\ul 0.865}  & 0.867       & 0.88      & 0.903   & 0.897  & 0.94    & 1.055    & 0.894    \\ \midrule
         & sMAPE     & \textbf{4.577}       & {\ul 4.742}  & 4.751       & 4.843     & 4.795   & 4.94   & 4.894   & 6.184    & 4.982    \\
Others   & MASE      & {\ul \textbf{3.117}} & {\ul 3.117}  & 3.127       & 3.277     & 3.178   & 3.228  & 3.358   & 4.818    & 3.323    \\
         & OWA       & \textbf{0.984}       & {\ul 0.993}  & 0.994       & 1.026     & 1.006   & 1.029  & 1.044   & 1.14     & 1.048    \\ \midrule
         & sMAPE     & \textbf{11.41}       & {\ul 11.558} & 11.567      & 11.831    & 11.983  & 11.991 & 12.418  & 13.022   & 11.93    \\
AVERAGE  & MASE      & \textbf{1.559}       & {\ul 1.571}  & 1.573       & 1.585     & 1.595   & 1.6    & 1.656   & 1.814    & 1.597    \\
         & OWA       & \textbf{0.828}       & {\ul 0.837}  & 0.838       & 0.85      & 0.859   & 0.861  & 0.891   & 0.954    & 0.867    \\ \midrule
         & 1st Count & \textbf{12}                   & 0            & 0           & 0         & 0       & 0      & 0       & 0        & 0        \\ 
\bottomrule
\end{tabular}
\end{adjustbox}

\end{table*}

\subsection{Experimental Settings}
\noindent\textbf{Datasets.}
We benchmark {SCALER} on standard long-horizon multivariate forecasting datasets used by recent LLM-based
forecasters, following the TimeLLM evaluation protocol \cite{time-llm}. Specifically, we use the four ETT variants (ETTh1/ETTh2/ETTm1/ETTm2) together with Electricity (ECL), Traffic, Weather, and ILI. For ETT/Weather /ECL/Traffic, we evaluate forecasting horizons $H\in\{96,192,336,720\}$, while for ILI we follow $H\in\{24,36,48,60\}$. We rely on the official training/validation/test partitions and the unified data processing pipeline provided by Time-Series-Library to keep the experimental setup consistent across all methods. Each variable is normalised using statistics computed on the training split. For long-term forecasting, we report MSE and MAE. For short-term forecasting, we report sMAPE, MASE, and OWA, and for zero-shot forecasting we report sMAPE. All metrics are evaluated on the test split, and results are averaged over three runs with different random seeds. The details of datasets are reported in \textbf{Appendix \ref{sec:dataset_desc}}
\begin{table*}[t]
\centering
\caption{\textbf{Zero-shot forecasting results on M3 \cite{m3} and M4 \cite{m4}.} We follow the evaluation protocol of \cite{llm-lvicl} and report sMAPE (lower is better) under two transfer settings: M3$\rightarrow$M4 and M4$\rightarrow$M3, evaluated on four scenario groups (Year, Quarter, Month, Others) and their average. \textbf{Bold} indicates the best result and underline indicates the second best within each row. The last row reports the number of first-place results across all scenario-category pairs. We report sMAPE as the evaluation metric, and all numbers are averaged over five runs with different random seeds.}
\label{tab:zeroshot_results}
\setlength{\tabcolsep}{3.2pt}
\renewcommand{\arraystretch}{1.05}
\begin{adjustbox}{width=\textwidth}
\begin{tabular}{@{}ll|ccccccccc@{}}
\toprule
Scenario & Category  & SCALER (Ours)   & TimeReasoner & LVICL  & AutoTimes & FPT           & DLinear & PatchTST & TimesNet & FEDformer \\ \midrule
         & Year      & \textbf{14.892} & {\ul 15.037} & 15.11  & 15.71     & 16.42         & 17.43   & 15.99    & 18.75    & 16        \\
         & Quarter   & \textbf{9.141}  & {\ul 9.287}  & 9.45   & 9.35      & 10.13         & 9.74    & 9.62     & 12.26    & 9.48      \\
M3$\rightarrow$M4 & Month     & \textbf{13.592} & {\ul 13.866} & 13.98  & 14.06     & 14.1          & 15.65   & 14.71    & 14.01    & 15.12     \\
         & Others    & {\ul 5.205}     & 5.477        & 5.59   & 5.79      & \textbf{4.81} & 6.81    & 9.44     & 6.88     & 8.94      \\
         & Average   & \textbf{12.36}  & {\ul 12.4}   & 12.596 & 12.75     & 13.06         & 14.03   & 13.39    & 14.17    & 13.53     \\ \midrule
         & Year      & \textbf{13.488} & {\ul 13.567} & 13.721 & 13.728    & 13.74         & 14.193  & 13.966   & 15.655   & 13.887    \\
         & Quarter   & \textbf{10.504} & {\ul 10.66}  & 10.734 & 10.742    & 10.787        & 18.856  & 10.929   & 11.877   & 11.513    \\
M4$\rightarrow$M3, & Month     & \textbf{14.299} & {\ul 14.498} & 14.56  & 14.558    & 14.63         & 14.765  & 14.664   & 16.165   & 18.154    \\
         & Others    & \textbf{5.912}  & {\ul 6.089}  & 6.219  & 6.259     & 7.081         & 9.194   & 7.087    & 6.863    & 7.529     \\
         & Average   & \textbf{12.76}  & {\ul 12.975} & 13.032 & 13.036    & 13.125        & 15.337  & 13.228   & 14.553   & 15.047    \\ \midrule
         & 1st Count & \textbf{9}      & 0            & 0      & 0         & 1             & 0       & 0        & 0        & 0         \\ \bottomrule
\end{tabular}
\end{adjustbox}

\end{table*}

\noindent\textbf{Baselines.}
We compare {SCALER} with two groups of representative methods, covering both standard deep forecasters and LLM-based forecasting approaches.

\noindent\textbf{(i) Standard forecasters.}
We include strong neural forecasting backbones that are widely used on these benchmarks: iTransformer, DLinear, PatchTST, and TimesNet.

\noindent\textbf{(ii) LLM-based forecasters and test-time scaling.}
We further compare against recent LLM-based and foundation-model forecasters: LVICL \cite{llm-lvicl}, AutoTimes \cite{llm-autotimes}, TimeLLM \cite{time-llm}, FPT \cite{llm-fpt}, Chronos \cite{chronos}, and TimesFM \cite{llm-timesfm}. We further compare against the recent test-time scaling for LLM-based time series forecasting: TimeReasoner \cite{timereasoner}. All baselines are evaluated with the same data splits, preprocessing, and metrics to ensure fair comparison. When a method involves test-time scaling, we additionally control the inference budget and keep the prediction horizon matched across methods.

\noindent\textbf{Implementation Details.}
We consider each channel separately, and fixed $\lambda$ and $\beta$ to 1, patch length $\ell = 12$, and set $\mathrm{Scale} \in {T/8,, T/4,, T/2,, T}$ for all experiments. All models are implemented in PyTorch and run on NVIDIA A100-80GB GPUs. We optimise with Adam and choose the learning rate from
$\{3\mathrm{e}{-}5, 5\mathrm{e}{-}5, 7\mathrm{e}{-}5, 9\mathrm{e}{-}5\}$, using MSE as the training objective. Following common practice for multivariate forecasting, we process each channel as an individual series in training and evaluation. We use the standard input lengths for these benchmarks (e.g., 672 for ETT/Weather/ECL/Traffic). The batch size is selected from $\{192,256,1024,2048\}$. We train for at most 40 epochs and apply early stopping when the validation loss does not improve for three consecutive evaluations.

\subsection{Long-Term Forecasting}

Table~\ref{tab:scaler_full_ltsf} reports long-term forecasting results averaged over four prediction horizons on seven benchmarks. {SCALER} achieves the best overall accuracy on most datasets, ranking first on \textbf{ETTh1}, \textbf{ETTh2}, \textbf{ETTm2}, \textbf{Weather} (MSE), and \textbf{Traffic}, and remaining highly competitive on \textbf{ECL} where it ties for the best MSE. Compared with LLM-based forecasters (e.g., LVICL, AutoTimes, TimeLLM, FPT, Chronos, and TimeFM), the proposed coarse-to-fine test-time scaling consistently reduces error while keeping the model efficient. The \textit{ 1st Count} row further summarises this trend, showing that {SCALER} obtains the largest number of first-place results across all dataset--horizon pairs, highlighting its robustness across diverse domains and forecasting lengths. Full results are reported in \textbf{Appendix~\ref{sec:full_results}}.

\subsection{Short-Term Forecasting}

Table~\ref{tab:stf_results} summarises short-term forecasting performance under four scenario groups (Year, Quarter, Month, and Others) using standard M4-style metrics (sMAPE, MASE, and OWA), where lower is better. {SCALER} consistently achieves the best results across all scenarios and metrics, obtaining the lowest sMAPE and MASE and therefore the lowest OWA in each group. In particular, {SCALER} improves over the strongest baseline, TimeReasoner, on every scenario, and yields clear margins over LLM-based forecasters (LVICL, AutoTimes, TimeLLM, and FPT) as well as competitive deep forecasting models (DLinear, PatchTST, and TimesNet). Averaged across scenarios, {SCALER} attains the best overall accuracy (sMAPE $=11.41$, MASE $=1.559$, OWA $=0.828$). The \textit{1st Count} row further confirms this robustness: {SCALER} ranks first in all 12 scenario--metric combinations, indicating that the proposed test-time scaling strategy generalises well across different temporal granularities and data characteristics.

\subsection{Zero-shot Forecasting}

\label{sec:zeroshot}
To examine cross-dataset generalisation without any target-domain adaptation, we further evaluate {SCALER} in a zero-shot setting on the \textbf{M3} \cite{m3} and \textbf{M4} \cite{m4} benchmarks, following the protocol in \cite{llm-lvicl}. Concretely, we consider two transfer directions: \textbf{M4$\rightarrow$M3} (train on M4 and test on M3) and \textbf{M3$\rightarrow$M4} (train on M3 and test on M4). For M4$\rightarrow$M3, Yearly/Quarterly/Monthly subsets are transferred from the corresponding M4 frequency to the matching M3 frequency, while the M3-Others subset is trained using M4-Quarterly. For M3$\rightarrow$M4, we similarly transfer across matched frequencies, and train M4-Others using M3-Monthly. We report sMAPE as the evaluation metric, and all numbers are averaged over five runs with different random seeds, consistent with \cite{llm-lvicl}.

Table~\ref{tab:zeroshot_results} shows that {SCALER} consistently achieves the best performance in both transfer settings. For M3$\rightarrow$M4, it obtains the lowest average sMAPE (12.36), surpassing TimeReasoner (12.40) and LVICL (12.596). For M4$\rightarrow$M3, it again ranks first across all categories with an average sMAPE of 12.76, outperforming TimeReasoner (12.975) and LVICL (13.032). The \textit{1st Count} metric further highlights its robustness, with {SCALER} achieving 9 first-place results, demonstrating effective test-time scaling across datasets and sampling frequencies without retraining.

\begin{figure*}[t]
    \centering
    \begin{adjustbox}{max width=\linewidth}
        \includegraphics{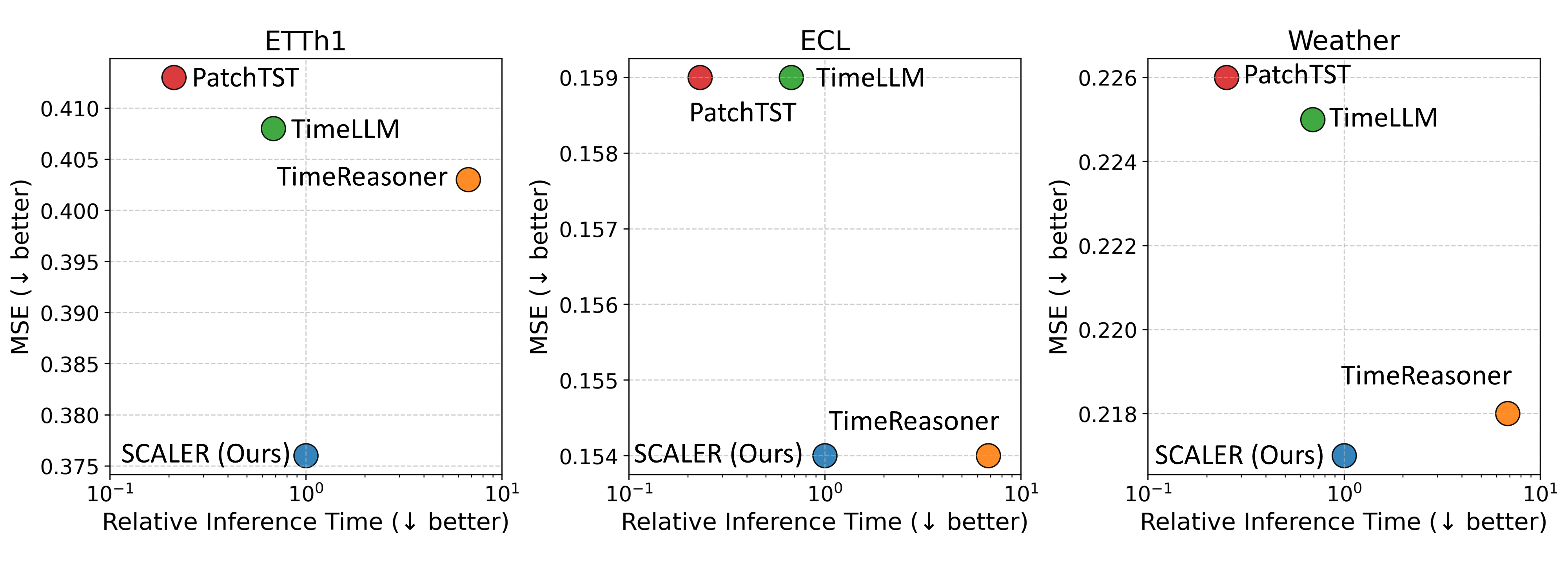}
    \end{adjustbox}
    \caption{\textbf{Accuracy--efficiency trade-off.}
    MSE (y-axis, lower is better) versus relative inference time (x-axis, lower is faster) on ETTh1, ECL, and Weather.
    {SCALER} consistently achieves the lowest (or tied-lowest) MSE while remaining computationally efficient.
    It is much faster than {TimeReasoner} (about $6.7\times$--$6.8\times$ under the same normalisation) and yields better accuracy, while also improving over {TimeLLM} and {PatchTST} in MSE with only modest extra inference cost.}
    \label{fig:time}
    
\end{figure*}

\subsection{Computation Time Comparison}
\label{sec:time}

We compare inference efficiency together with forecasting accuracy on three standard long-term TSF benchmarks (ETTh1, ECL, and Weather).
Following common practice, we report MSE (lower is better) and relative inference time (lower is faster), where {SCALER} is used as the runtime reference and normalised to $1\times$ on each dataset.

Figure~\ref{fig:time} summarises the accuracy--efficiency trade-off by plotting MSE against relative inference time for all methods.
Across all datasets, {SCALER} consistently achieves the lowest (or tied-lowest) MSE while remaining computationally efficient.
In particular, {SCALER} is substantially faster than {TimeReasoner} (about $6.7\times$--$6.8\times$ under the same normalisation) and also yields better accuracy, indicating that heavy test-time scaling can be avoided without sacrificing performance.
Compared with faster baselines such as {TimeLLM} and {PatchTST}, {SCALER} provides a clear accuracy gain (lower MSE) at a modest additional inference cost.
Overall, these results place {SCALER} on a favourable Pareto frontier, demonstrating improved forecasting quality without compromising practical inference efficiency.


\subsection{Ablation Studies}

\subsubsection{Global Forecaster.}
We evaluate the benefit of predicting global structure before refinement under two settings:
\begin{itemize}[leftmargin=*, nosep]
    \item \textbf{(i) w/o Global Forecaster:} remove $\mathcal{F}$ and feed only the history tokens $Z_I$ to the pretrained LLM refiner.
    \item \textbf{(ii) Global Forecaster (Full Tokens):} use $\mathcal{F}$ to predict full-resolution future tokens, then refine them with the same pretrained LLM.
\end{itemize}
As shown in the first two method groups of Table~\ref{tab:ab1}, introducing a global forecaster consistently reduces both MSE and MAE across datasets and horizons, confirming that explicit global-shape modeling is beneficial. However, forecasting \emph{full-resolution} tokens is substantially more expensive and does not yield better accuracy. We attribute this to the role of the global forecaster: it models low-frequency structure well, but its fine-grained predictions can be unreliable, which may mislead the LLM during refinement. In contrast, {SCALER} predicts only a downscaled coarse shape and applies fixed-step refinement, achieving the best (or near-best) accuracy with much lower cost.

\begin{table*}[]
\centering
\caption{Ablation analysis for our component choices. We report Inference and Training Time normalised by SCALER (SCALER = 1.0).}
\label{tab:ab1}
\centering
\setlength{\tabcolsep}{3.2pt}
\renewcommand{\arraystretch}{1.05}
\begin{adjustbox}{width=\textwidth}
\begin{tabular}{@{}ll|cc|cc|cc|cc|cc|cc|cc|cc@{}}
\toprule
        &                    & \multicolumn{4}{c}{Global   Forecaster Analysis}                                              & \multicolumn{4}{|c}{Scaling   Technique Analysis}                              & \multicolumn{6}{|c}{Tuning   Technique Analysis}      &                                                     \multicolumn{2}{|c}{Default}                             \\ \midrule
        &                    & \multicolumn{2}{c}{w/o Global Forecaster} & \multicolumn{2}{|c}{Global Forecaster (Full)} & \multicolumn{2}{|c}{w/ Description Tokens} & \multicolumn{2}{|c}{w/ Reward Model} & \multicolumn{2}{|c}{w/ Prompt Tuning} & \multicolumn{2}{|c}{w/ LoRA Finetuning} & \multicolumn{2}{|c}{w/ Full Finetuning} & \multicolumn{2}{|c}{SCALER (Default)} \\ \midrule
Dataset & Length             & MSE                 & MAE                 & MSE                           & MAE                 & MSE                 & MAE                 & MSE                   & MAE         & MSE                    & MAE                    & MSE                    & MAE           & MSE                & MAE               & MSE                & MAE             \\ \midrule
ETTh1   & 96                 & 0.367               & 0.401               & 0.349                      & 0.388                  & 0.35                & 0.389               & 0.354                 & 0.39        & 0.349                  & 0.389                  & 0.347               & {\ul 0.387}      & 0.346              & {\ul 0.387}       & \textbf{0.345}    & \textbf{0.386}   \\
        & 192                & 0.408               & 0.423               & 0.381                      & 0.41                   & 0.378               & 0.411               & 0.381                 & 0.413       & 0.378                  & 0.41                   & 0.376               & 0.408            & 0.375              & {\ul 0.407}       & \textbf{0.374}    & \textbf{0.406}   \\
        & 336                & 0.417               & 0.427               & 0.389                      & 0.417                  & 0.39                & 0.416               & 0.392                 & 0.419       & 0.389                  & 0.416                  & 0.387               & 0.415            & 0.386              & {\ul 0.414}       & \textbf{0.385}    & \textbf{0.413}   \\
        & 720                & 0.434               & 0.453               & 0.405                      & 0.429                  & 0.404               & 0.429               & 0.406                 & 0.428       & 0.403                  & 0.429                  & 0.401               & 0.428            & 0.4                & {\ul 0.426}       & \textbf{0.399}    & \textbf{0.425}   \\
        & Avg                & 0.407               & 0.426               & 0.381                      & 0.411                  & 0.381               & 0.411               & 0.383                 & 0.413       & 0.38                   & 0.411                  & 0.378               & 0.41             & \textbf{0.375}     & \textbf{0.406}    & 0.376             & {\ul 0.408}      \\ \midrule
ECL     & 96                 & 0.147               & 0.231               & 0.125                      & 0.225                  & 0.126               & 0.226               & 0.127                 & 0.225       & 0.124                  & 0.225                  & 0.122               & {\ul 0.223}      & 0.121              & {\ul 0.223}       & \textbf{0.12}     & \textbf{0.222}   \\
        & 192                & 0.175               & 0.246               & 0.146                      & 0.233                  & 0.149               & 0.235               & 0.152                 & 0.235       & 0.148                  & 0.235                  & 0.146               & 0.233            & 0.144              & {\ul 0.232}       & \textbf{0.143}    & \textbf{0.231}   \\
        & 336                & 0.188               & 0.252               & 0.165                      & 0.248                  & 0.163               & 0.251               & 0.166                 & 0.25        & 0.161                  & 0.25                   & 0.16                & 0.249            & \textbf{0.156}     & \textbf{0.245}    & 0.158             & {\ul 0.246}      \\
        & 720                & 0.223               & 0.29                & 0.199                      & 0.284                  & 0.2                 & 0.285               & 0.201                 & 0.286       & 0.198                  & 0.284                  & 0.196               & 0.283            & 0.195              & {\ul 0.282}       & \textbf{0.194}    & \textbf{0.281}   \\
        & Avg                & 0.183               & 0.255               & 0.159                      & 0.248                  & 0.16                & 0.249               & 0.162                 & 0.249       & 0.158                  & 0.249                  & 0.156               & 0.247            & \textbf{0.152}     & \textbf{0.244}    & 0.154             & {\ul 0.245}      \\ \midrule
Weather & 96                 & 0.162               & 0.212               & 0.145                      & 0.196                  & 0.146               & 0.197               & 0.149                 & 0.196       & 0.144                  & 0.197                  & 0.143               & 0.196            & 0.142              & {\ul 0.195}       & \textbf{0.141}    & \textbf{0.193}   \\
        & 192                & 0.22                & 0.262               & 0.19                       & 0.241                  & 0.191               & 0.241               & 0.191                 & 0.242       & 0.189                  & 0.241                  & 0.187               & {\ul 0.239}      & 0.187              & {\ul 0.239}       & \textbf{0.186}    & \textbf{0.238}   \\
        & 336                & 0.252               & 0.301               & 0.24                       & 0.281                  & 0.239               & 0.28                & 0.242                 & 0.279       & 0.238                  & 0.279                  & 0.236               & {\ul 0.278}      & 0.235              & {\ul 0.278}       & \textbf{0.234}    & \textbf{0.277}   \\
        & 720                & 0.329               & 0.346               & 0.31                       & 0.323                  & 0.311               & 0.322               & 0.314                 & 0.324       & 0.309                  & 0.324                  & 0.308               & 0.322            & \textbf{0.305}     & {\ul 0.321}       & 0.306             & \textbf{0.32}    \\
        & Avg                & 0.241               & 0.28                & 0.221                      & 0.26                   & 0.222               & 0.26                & 0.224                 & 0.26        & 0.22                   & 0.26                   & 0.219               & 0.259            & 0.218              & {\ul 0.258}       & \textbf{0.217}    & \textbf{0.257}   \\ \midrule
Traffic & 96                 & 0.362               & 0.258               & 0.339                      & {\ul 0.246}            & 0.34                & 0.249               & 0.34                  & 0.25        & 0.337                  & 0.248                  & 0.336               & 0.247            & 0.335              & {\ul 0.246}       & \textbf{0.334}    & \textbf{0.244}   \\
        & 192                & 0.389               & 0.273               & 0.361                      & 0.252                  & 0.363               & 0.254               & 0.364                 & 0.257       & 0.361                  & 0.253                  & 0.36                & 0.252            & 0.359              & {\ul 0.251}       & \textbf{0.358}    & \textbf{0.25}    \\
        & 336                & 0.403               & 0.27                & 0.381                      & 0.258                  & 0.383               & 0.256               & 0.384                 & 0.256       & 0.382                  & 0.258                  & 0.38                & 0.256            & \textbf{0.378}     & \textbf{0.253}    & \textbf{0.378}    & {\ul 0.254}      \\
        & 720                & 0.423               & 0.286               & 0.404                      & 0.28                   & 0.403               & 0.281               & 0.405                 & 0.279       & 0.403                  & 0.279                  & 0.401               & 0.278            & 0.4                & {\ul 0.277}       & \textbf{0.399}    & \textbf{0.276}   \\
        & Avg                & 0.394               & 0.272               & 0.371                      & 0.259                  & 0.372               & 0.26                & 0.373                 & 0.261       & 0.371                  & 0.26                   & 0.369               & 0.258            & 0.368              & {\ul 0.257}       & \textbf{0.367}    & \textbf{0.256} \\ \midrule
        \multicolumn{2}{c}{Inference Time}  & \multicolumn{2}{|c}{0.956}                 & \multicolumn{2}{|c}{1.841}                           & \multicolumn{2}{|c}{2.341}                 & \multicolumn{2}{|c}{5.512}           & \multicolumn{2}{|c}{1.052}                       & \multicolumn{2}{|c}{1}                  & \multicolumn{2}{|c}{1}                  & \multicolumn{2}{|c}{1}                \\ \midrule
        \multicolumn{2}{|c}{Training Time}  & \multicolumn{2}{c}{0.912}                 & \multicolumn{2}{|c}{1.724}                           & \multicolumn{2}{|c}{2.248}                 & \multicolumn{2}{|c}{5.612}           & \multicolumn{2}{|c}{1.211}                       & \multicolumn{2}{|c}{8.21}               & \multicolumn{2}{|c}{15.21}              & \multicolumn{2}{|c}{1}                \\ \bottomrule
\end{tabular}
\end{adjustbox}
\end{table*}

\subsubsection{Scaling Techniques.}
Next, we study scaling-time inference strategies that aim to improve accuracy by increasing test-time compute:
\begin{itemize}[leftmargin=*, nosep]
    \item \textbf{w/ Description Tokens:} augment the LLM input with additional natural-language description tokens (following prior LLM-forecasting protocols) to provide extra guidance.
    \item \textbf{(w/ Reward Model:} generate multiple candidates and use a reward model to select the best prediction, further increasing inference cost.
\end{itemize}
The middle method groups in Table~\ref{tab:ab1} show that these strategies provide only marginal (and
sometimes inconsistent) gains over the global-forecaster baseline, while noticeably increasing inference
time (especially reward-model selection). In contrast, {SCALER} uses fixed-step refinement and
avoids candidate generation/selection, achieving better accuracy with predictable and efficient scaling.

\subsubsection{Tuning Techniques.}
Finally, we compare common parameter-efficient and full-parameter tuning strategies for adapting the
pretrained LLM to forecasting:
\begin{itemize}[leftmargin=*, nosep]
    \item \textbf{(w/ Prompt Tuning:} learn a small set of soft prompt embeddings while keeping the LLM frozen.
    \item \textbf{(w/ LoRA Finetuning:} insert low-rank adapters and update only adapter parameters.
    \item \textbf{(w/ Full Finetuning:} update all model parameters end-to-end.
\end{itemize}
As shown in the corresponding method groups of Table~\ref{tab:ab1}, tuning can improve performance over
pure prompting-based baselines, but it comes with a clear training-time overhead (and full finetuning is
the most expensive). In contrast, {SCALER} achieves the best (or near-best) accuracy without heavy
tuning by using a downscaled coarse-shape prior and fixed-step refinement, offering a more efficient
accuracy, cost trade-off.

Further ablation studies are reported in \textbf{Appendix \ref{sec:further_as}}.


\subsection{Further Analysis}
\subsubsection{Additional Results for Different Global Forecaster Backbone Sizes}
\label{sec:global_backbone}

\begin{figure}[t]
    \centering
    \begin{adjustbox}{max width=\linewidth}
        \includegraphics{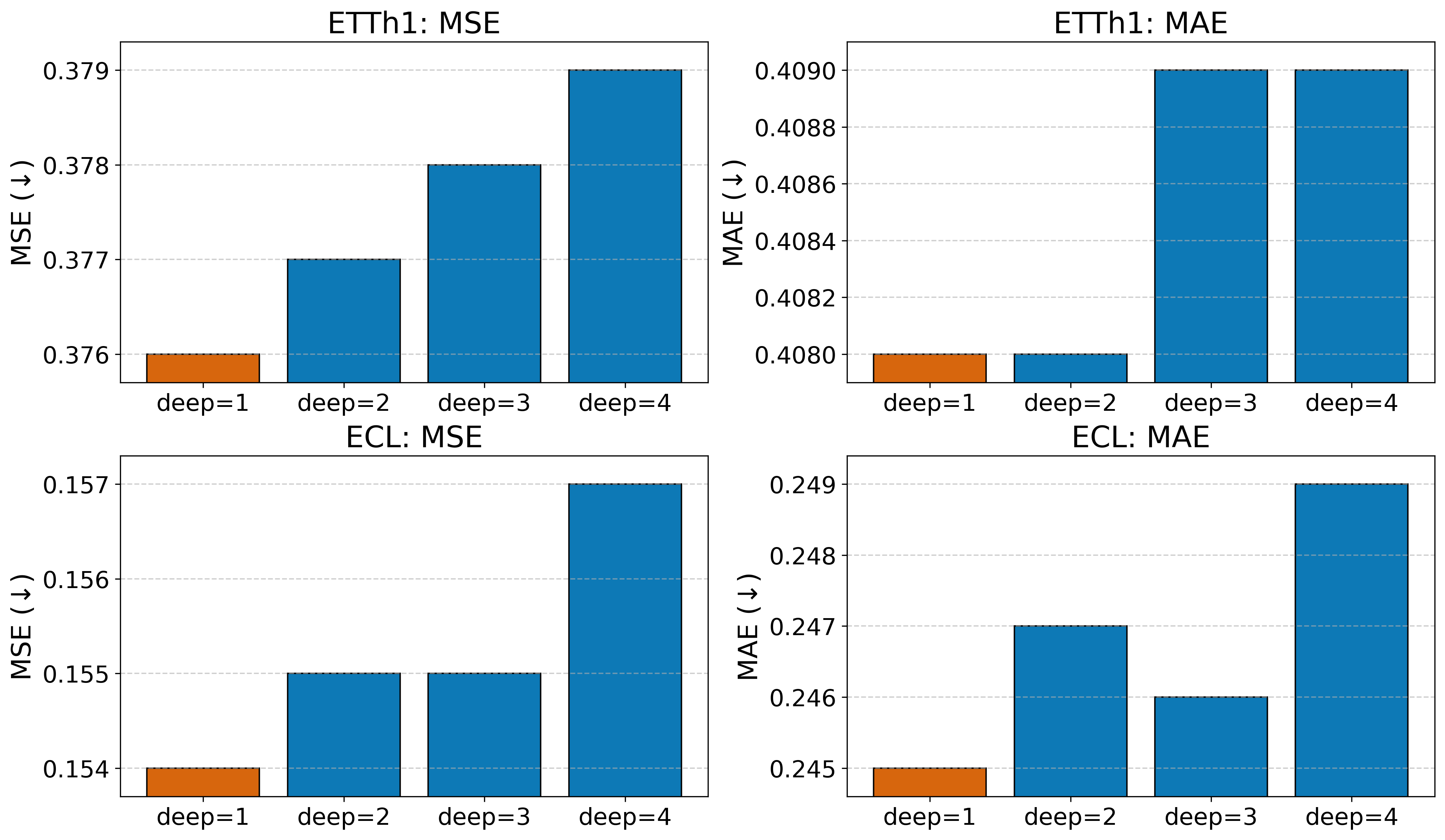}
    \end{adjustbox}
    \caption{\textbf{Effect of the depth of global forecaster.}
    We vary the depth of the lightweight global forecaster from $1$ to $4$ layers while keeping the hidden dimension $d_{\text{model}}$ fixed to match the LLM backbone.
    The first setting (depth$=1$, orange) is our default configuration, and deeper variants (blue) show only marginal changes in both MSE and MAE on ETTh1 and ECL.}
    \label{fig:depth_ablation}
    
\end{figure}

To study the impact of global-forecaster capacity, we fix its hidden size to match the LLM backbone (same $d_{\text{model}}$) and vary only the depth from 1 to 4 layers. Figure~\ref{fig:depth_ablation} shows the MSE and MAE on ETTh1 and ECL. Performance is largely stable across depths: deeper models bring only minor, inconsistent gains. Notably, depth$=1$ already achieves the best or near-best results, suggesting that matching width (aligned with the LLM) is sufficient to provide an effective coarse shape, while extra depth offers diminishing returns. This supports the use of a lightweight global forecaster for efficient inference without compromising accuracy.

\subsubsection{Additional Results on Different LLM Backbones}
\label{sec:llm_backbone}

To test backbone sensitivity, we select the Stage~II LLM from LLaMA-7B (default), OPT-2B, OPT-1.3B, OPT-350M, and GPT-2 (124M), while keeping the global forecaster and the refinement pipeline unchanged.
Table~\ref{tab:llm_backbone} shows that {SCALER} remains effective across different families and scales of LLMs: smaller backbones are still competitive, while larger ones consistently improve accuracy.
In particular, LLaMA-7B achieves the best performance on both datasets (ETTh1: 0.376 MSE / 0.408 MAE; ECL: 0.154 MSE / 0.245 MAE), indicating that our framework generalises well and supports an accuracy--efficiency trade-off via backbone choice.

\begin{table}[t]

\centering
\setlength{\tabcolsep}{6pt}
\small
\begin{adjustbox}{max width=\linewidth}
\begin{tabular}{lccccc}
\toprule
\textbf{Dataset} & \textbf{LLaMA-7B (default)} & \textbf{OPT-2B} & \textbf{OPT-1.3B} & \textbf{OPT-350M} & \textbf{GPT-2 (124M)} \\
\midrule
ETTh1 (MSE$\downarrow$) & \textbf{0.376} & 0.379 & 0.381 & 0.381 & 0.379 \\
ETTh1 (MAE$\downarrow$) & \textbf{0.408} & 0.412 & 0.414 & 0.414 & 0.411 \\
\midrule
ECL (MSE$\downarrow$) & \textbf{0.154} & 0.159 & 0.160 & 0.161 & 0.159 \\
ECL (MAE$\downarrow$) & \textbf{0.245} & 0.253 & 0.256 & 0.259 & 0.254 \\
\bottomrule
\end{tabular}
\end{adjustbox}
\caption{\textbf{Effect of different LLM backbones on Stage~II refinement.}
{SCALER} remains effective across a wide range of sizes and families of LLMs, while larger backbones (e.g., LLaMA-7B) consistently yield higher accuracy on ETTh1 and ECL.}
\label{tab:llm_backbone}
\end{table}

\subsubsection{Performance Under Longer Forecasting Horizons}
\label{sec:longer_horizon}

\begin{figure}[t]
    \centering
    \begin{adjustbox}{max width=\linewidth}
        \includegraphics{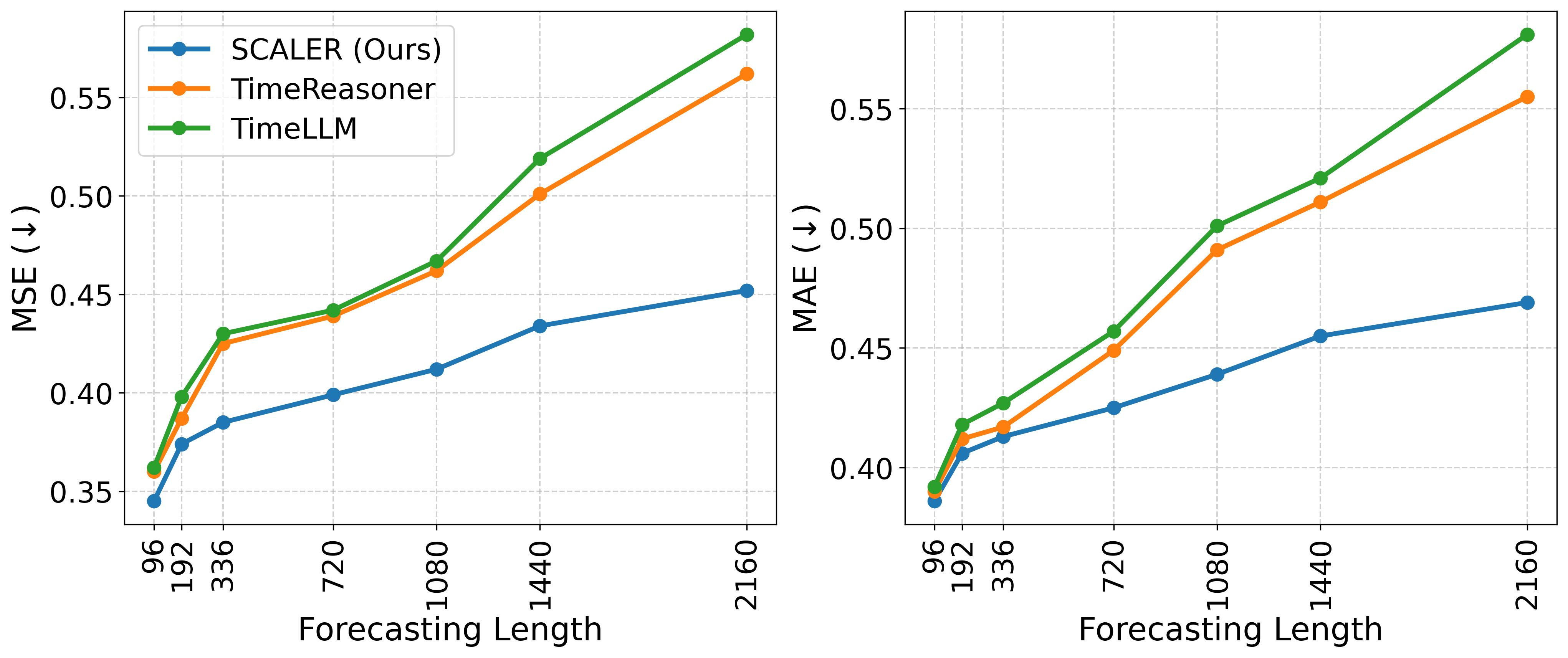}
    \end{adjustbox}
    \caption{\textbf{Long-horizon robustness on ETTh1.}
    MSE/MAE versus forecasting horizon $H\in\{96,192,336,720,1080,1440,2160\}$.
    {SCALER} degrades more gracefully as the horizon increases, and the performance gap over
    {TimeReasoner} and {TimeLLM} grows at longer horizons.}
    \label{fig:horizon_curve}
    \vspace{-6pt}
\end{figure}

To further examine robustness as the forecasting horizon increases beyond standard settings, we compare
{SCALER} with {TimeReasoner} and {TimeLLM} on progressively longer horizons and report
both MSE and MAE. Figure~\ref{fig:horizon_curve} shows a clear trend: while all methods incur larger
errors as the horizon grows, {SCALER} degrades much more gracefully and the advantage becomes more
pronounced at longer horizons. In particular, the gap between {SCALER} and the baselines widens
steadily for $H\ge 720$, suggesting that methods without an explicit global-shape anchor are increasingly
prone to accumulated mismatch in the long range. In contrast, {SCALER} maintains a more stable
trajectory by first predicting a coarse future shape and then refining details in a coarse-to-fine
manner, which better preserves long-term structure. Overall, these results highlight that the proposed
shape-anchored refinement is especially beneficial when the forecasting length becomes large, where
preserving global dynamics is critical and local-only refinements tend to compound errors.

\subsection{Shared Encoder Design and Training Stability}
 
SCALER uses a \emph{single shared encoder} $\mathcal{E}$ to tokenise both the input history
and the downscaled ground-truth targets at each scale.
This design choice is motivated by representation alignment:
using a shared encoder ensures that the history tokens $Z_I$ and the scale-$k$ target tokens
$Z_{S^k}$ live in the same embedding space,
which avoids cross-space mismatch and stabilises training.
 
While the shared encoder is jointly optimised with the rest of the framework,
the multi-task supervision structure---combining
$\mathcal{L}_{\text{shape}}$, $\mathcal{L}_{\text{refine}}$, and $\mathcal{L}_{\text{ts}}$---provides
complementary gradient signals that collectively prevent overfitting to any single scale.
The sensitivity study in Appendix~\ref{sec:further_as}
(Table~\ref{tab:ab_lambdabeta_full}) confirms that SCALER is robust to a wide range of
loss weight combinations $(\lambda, \beta) \in [0.5, 2]^2$,
which is consistent with stable training dynamics.
The consistent performance gains reported throughout Section~4 further
support the effectiveness of this design.

\section{Visualization of Global Shape Drift Mitigation}

\begin{figure}[htbp]
    \centering
    \includegraphics[width=\linewidth]{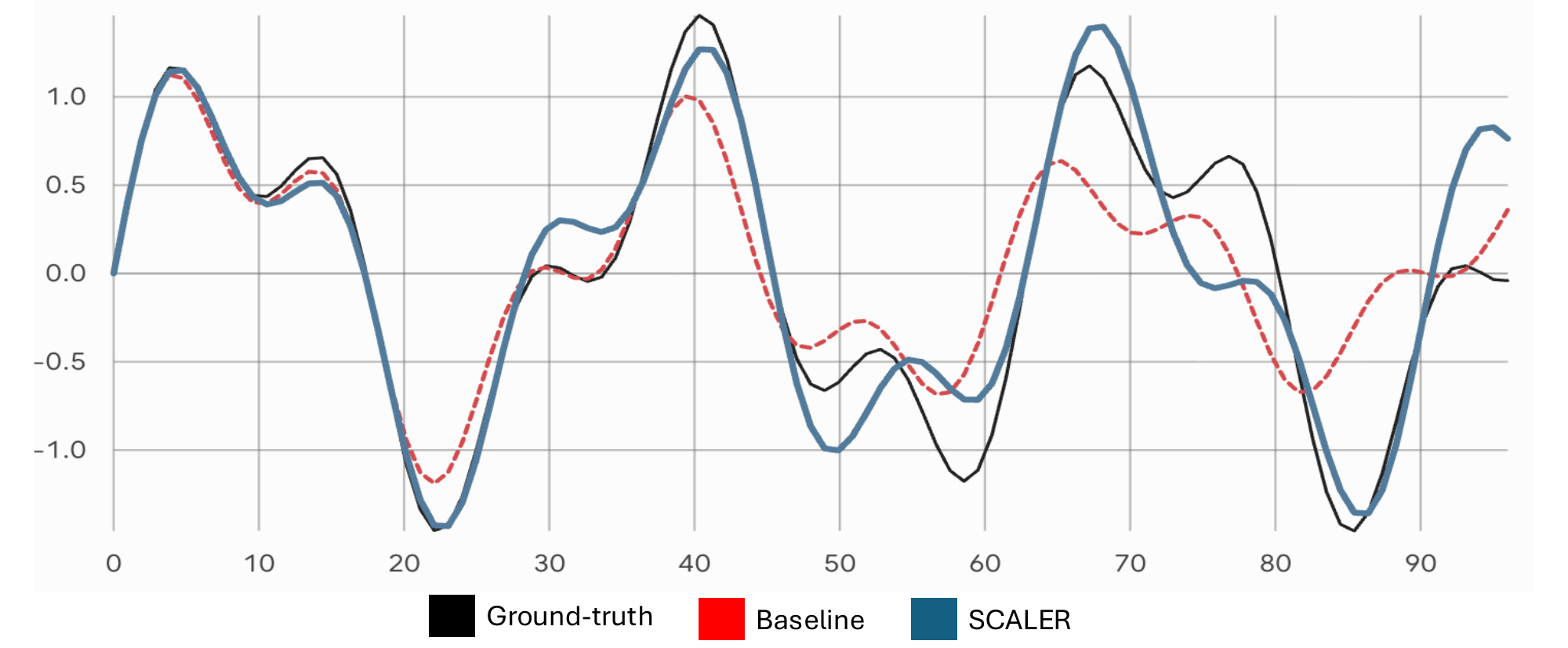}
    \caption{Representative forecast comparison. Black: ground truth; red dashed: baseline (global shape drift visible from step~40); blue: SCALER (maintains alignment throughout).}
    \label{fig:visual2}
\end{figure}

Figure~\ref{fig:visual2} shows a representative 96-step forecast comparing SCALER against a baseline. The baseline exhibits pronounced global shape drift from step 40 onward, underestimating peak magnitudes and failing to recover the correct oscillatory structure. SCALER closely tracks the ground truth throughout, faithfully reproducing peak locations, trough depths, and the amplitude envelope. This confirms that the proposed method produces structurally coherent predictions rather than merely improving aggregate error metrics.

\section{Conclusions, Limitations and Future Work}

We present \textsc{SCALER}, a coarse-to-fine framework that enables \emph{test-time scaling} for long-horizon multivariate forecasting with a pretrained LLM while preserving global structure. \textsc{SCALER} first predicts an explicit downscaled future shape using a lightweight forecaster, then allocates a fixed $K$-step refinement budget to an LLM refiner conditioned on this shape. This turns additional test-time compute into progressively finer token blocks, producing shape-faithful forecasts with predictable cost and without candidate sampling or reward-model selection. Experiments on standard TSF benchmarks show that \textsc{SCALER} consistently improves accuracy and delivers a stronger accuracy, efficiency trade-off than prior LLM-based test-time scaling approaches. 

\noindent\textbf{Limitations and Future Work.} \textsc{SCALER} currently relies on a fixed refinement schedule, and performance may depend on the LLM backbone and tokenisation strategy. Future work includes adaptive refinement policies, more efficient backbone models, and extensions to challenging real-world scenarios such as missing data and distribution shifts.

\bibliographystyle{ACM-Reference-Format}
\bibliography{sample-base}

\appendix
\section{Datasets}

\label{sec:dataset_desc}
We evaluate on diverse real-world forecasting benchmarks covering multiple domains and temporal resolutions. ETT contains electricity transformer measurements with four subsets (ETTh1/ETTh2 hourly, ETTm1/ETTm2 15-minute). Weather includes 21 meteorological variables sampled every 10 minutes, while ECL and Traffic contain hourly electricity consumption and road occupancy data, respectively. We also evaluate on the widely used univariate M3 \cite{m3} and M4 \cite{m4} benchmarks spanning multiple domains and frequencies. Following prior long-term forecasting studies, all datasets use the same preprocessing and chronological train/validation/test splits to prevent information leakage. Dataset statistics are summarised in Table~\ref{tab:dataset_detail}.

\begin{table}[h]

\centering
\caption{\textbf{Detailed dataset statistics.} }
\label{tab:dataset_detail}
\setlength{\tabcolsep}{3.2pt}
\renewcommand{\arraystretch}{1.05}
\begin{adjustbox}{width=\linewidth}
\begin{tabular}{l c c c c l}
\toprule
\textbf{Dataset} & \textbf{Dim} & \textbf{Prediction Length} & \textbf{Dataset Size (Tr/Val/Te)} & \textbf{Frequency} & \textbf{Domain} \\
\midrule
ETTh1, ETTh2     & 7   & \{96, 192, 336, 720\} & (8545, 2881, 2881)     & Hourly    & Electricity \\
ETTm1, ETTm2     & 7   & \{96, 192, 336, 720\} & (34465, 11521, 11521)  & 15 min    & Electricity \\
Weather          & 21  & \{96, 192, 336, 720\} & (36792, 5271, 10540)   & 10 min    & Weather \\
ECL              & 321 & \{96, 192, 336, 720\} & (18317, 2633, 5261)    & Hourly    & Electricity \\
Traffic          & 862 & \{96, 192, 336, 720\} & (12185, 1757, 3509)    & Hourly    & Transportation \\
\midrule
M4-Yearly        & 1   & 6   & (23000, 0, 23000) & Yearly    & Demographic \\
M4-Quarterly     & 1   & 8   & (24000, 0, 24000) & Quarterly & Finance \\
M4-Monthly       & 1   & 18  & (48000, 0, 48000) & Monthly   & Industry \\
M4-Weekly        & 1   & 13  & (359, 0, 359)     & Weekly    & Macro \\
M4-Daily         & 1   & 14  & (4227, 0, 4227)   & Daily     & Micro \\
M4-Hourly        & 1   & 48  & (414, 0, 414)     & Hourly    & Other \\
\midrule
M3-Yearly        & 1   & 6   & (645, 0, 645)     & Yearly    & Demographic \\
M3-Quarterly     & 1   & 8   & (756, 0, 756)     & Quarterly & Finance \\
M3-Monthly       & 1   & 18  & (1428, 0, 1428)   & Monthly   & Industry \\
M3-Others        & 1   & 8   & (174, 0, 174)     & Weekly    & Macro \\
\bottomrule
\end{tabular}
\end{adjustbox}

\end{table}
\section{Evaluation Metrics}
\label{sec:metrics}

Let $H$ denote the prediction horizon, and let $Y_h$ and $\hat{Y}_h$ be the ground-truth and
predicted value at step $h \in \{1,\ldots,H\}$. We compute:
\begin{align}
\mathrm{sMAPE} &= \frac{200}{H}\sum_{h=1}^{H}\frac{\left|Y_h-\hat{Y}_h\right|}{\left|Y_h\right|+\left|\hat{Y}_h\right|}, 
\end{align}

For MASE, we normalise the absolute error by the in-sample seasonal naive error:
\begin{equation}
\mathrm{MASE}
=
\frac{1}{H}\sum_{h=1}^{H}
\frac{\left|Y_h-\hat{Y}_h\right|}
{\frac{1}{H-s}\sum_{j=s+1}^{H}\left|Y_j-Y_{j-s}\right|},
\end{equation}
where $s$ is the seasonality (periodicity) of the series.

Finally, OWA combines the relative sMAPE and MASE compared to the M4 Naive2 baseline:
\begin{equation}
\mathrm{OWA}
=
\frac{1}{2}\left(
\frac{\mathrm{sMAPE}}{\mathrm{sMAPE}_{\text{Naive2}}}
+
\frac{\mathrm{MASE}}{\mathrm{MASE}_{\text{Naive2}}}
\right).
\end{equation}

\section{Further Ablation Studies}
\label{sec:further_as}

\noindent\textbf{Parameters $\lambda$ and $\beta$.}
We sweep $\lambda,\beta\in\{0.5,1,1.5,2\}$ on ETTh1 and ECL with all other settings fixed. As Table~\ref{tab:ab_lambdabeta_full} shows, results are stable across all combinations, and the default $(\lambda,\beta)=(1,1)$ achieves the best or tied-best performance (ETTh1: 0.376/0.408; ECL: 0.154/0.245 MSE/MAE).

\begin{table}[t]
\centering
\small
\setlength{\tabcolsep}{4pt}
\begin{minipage}{0.49\linewidth}
\centering
\begin{adjustbox}{max width=\linewidth}
\begin{tabular}{lcccc}
\toprule
\multicolumn{5}{c}{\textbf{ETTh1 MSE$\downarrow$}} \\
\midrule
 & $\beta{=}0.5$ & $\beta{=}1$ & $\beta{=}1.5$ & $\beta{=}2$ \\
\midrule
$\lambda{=}0.5$ & 0.378 & 0.377 & 0.379 & 0.381 \\
$\lambda{=}1$   & \textbf{0.377} & \textbf{0.376} & 0.377 & 0.381 \\
$\lambda{=}1.5$ & 0.378 & 0.377 & 0.379 & 0.379 \\
$\lambda{=}2$   & 0.379 & 0.381 & 0.378 & 0.379 \\
\bottomrule
\end{tabular}
\end{adjustbox}
\end{minipage}
\hfill
\begin{minipage}{0.49\linewidth}
\centering
\begin{adjustbox}{max width=\linewidth}
\begin{tabular}{lcccc}
\toprule
\multicolumn{5}{c}{\textbf{ETTh1 MAE$\downarrow$}} \\
\midrule
 & $\beta{=}0.5$ & $\beta{=}1$ & $\beta{=}1.5$ & $\beta{=}2$ \\
\midrule
$\lambda{=}0.5$ & 0.411 & 0.410 & 0.410 & 0.411 \\
$\lambda{=}1$   & 0.409 & \textbf{0.408} & 0.410 & 0.411 \\
$\lambda{=}1.5$ & 0.409 & 0.409 & \textbf{0.408} & 0.410 \\
$\lambda{=}2$   & 0.411 & 0.409 & 0.409 & 0.411 \\
\bottomrule
\end{tabular}
\end{adjustbox}
\end{minipage}
\vspace{3pt}
\begin{minipage}{0.49\linewidth}
\centering
\begin{adjustbox}{max width=\linewidth}
\begin{tabular}{lcccc}
\toprule
\multicolumn{5}{c}{\textbf{ECL MSE$\downarrow$}} \\
\midrule
 & $\beta{=}0.5$ & $\beta{=}1$ & $\beta{=}1.5$ & $\beta{=}2$ \\
\midrule
$\lambda{=}0.5$ & 0.156 & \textbf{0.154} & 0.155 & 0.158 \\
$\lambda{=}1$   & 0.156 & \textbf{0.154} & 0.156 & 0.158 \\
$\lambda{=}1.5$ & 0.155 & 0.155 & 0.155 & 0.157 \\
$\lambda{=}2$   & 0.158 & 0.157 & 0.157 & 0.158 \\
\bottomrule
\end{tabular}
\end{adjustbox}
\end{minipage}
\hfill
\begin{minipage}{0.49\linewidth}
\centering
\begin{adjustbox}{max width=\linewidth}
\begin{tabular}{lcccc}
\toprule
\multicolumn{5}{c}{\textbf{ECL MAE$\downarrow$}} \\
\midrule
 & $\beta{=}0.5$ & $\beta{=}1$ & $\beta{=}1.5$ & $\beta{=}2$ \\
\midrule
$\lambda{=}0.5$ & 0.246 & 0.246 & 0.247 & 0.248 \\
$\lambda{=}1$   & 0.246 & \textbf{0.245} & 0.248 & 0.247 \\
$\lambda{=}1.5$ & \textbf{0.245} & 0.247 & 0.246 & 0.248 \\
$\lambda{=}2$   & 0.248 & 0.247 & 0.248 & 0.248 \\
\bottomrule
\end{tabular}
\end{adjustbox}
\end{minipage}
\caption{\textbf{Sensitivity to $\lambda$ and $\beta$ (MSE/MAE$\downarrow$).} Rows vary $\lambda$, columns vary $\beta$, in $\{0.5,1,1.5,2\}$.}
\label{tab:ab_lambdabeta_full}
\end{table}

\noindent\textbf{Patch length $\ell$.}
Table~\ref{tab:ab_patchlen} reports results for $\ell\in\{6,12,16,24\}$, each paired with an appropriate scale schedule. Performance is stable for smaller patches, but degrades with coarser tokenisation ($\ell{\ge}16$), likely due to loss of fine-grained temporal detail. We adopt $\ell{=}12$ as the default.

\begin{table}[hpt]
\centering
\small
\setlength{\tabcolsep}{6pt}
\begin{adjustbox}{max width=0.65\linewidth}
\begin{tabular}{lcccc}
\toprule
\textbf{Dataset} & $\boldsymbol{\ell{=}6}$ & $\boldsymbol{\ell{=}12}$ & $\boldsymbol{\ell{=}16}$ & $\boldsymbol{\ell{=}24}$ \\
\midrule
ETTh1 (MSE$\downarrow$) & 0.377 & \textbf{0.376} & 0.379 & 0.381 \\
ETTh1 (MAE$\downarrow$) & 0.409 & \textbf{0.408} & 0.412 & 0.414 \\
\midrule
ECL (MSE$\downarrow$) & 0.156 & \textbf{0.154} & 0.158 & 0.161 \\
ECL (MAE$\downarrow$) & 0.247 & \textbf{0.245} & 0.248 & 0.259 \\
\bottomrule
\end{tabular}
\end{adjustbox}
\caption{\textbf{Patch length ablation.} Scale schedules: $\ell{=}6\!:\{1,2,4,8\}$, $\ell{=}12\!:\{2,4,8,16\}$, $\ell{=}16\!:\{1,2,4,6\}$, $\ell{=}24\!:\{1,2,3,4\}$.}
\label{tab:ab_patchlen}
\end{table}

\noindent\textbf{Scale Token Schedule.}
Table~\ref{tab:ab_scale_list} evaluates schedules with 1--5 refinement steps. Accuracy improves consistently up to 4 steps, with gains saturating beyond that. The 4-step schedule $\{1,2,4,8\}$ achieves the best results on both datasets and is used as default; extending to 5 steps offers no further benefit.

\begin{table}[t]
\centering
\small
\setlength{\tabcolsep}{6pt}
\begin{adjustbox}{max width=0.9\linewidth}
\begin{tabular}{l l c c c c}
\toprule
\textbf{Steps} & \textbf{Scale List} & \textbf{ETTh1-MSE$\downarrow$} & \textbf{ETTh1-MAE$\downarrow$} & \textbf{ECL-MSE$\downarrow$} & \textbf{ECL-MAE$\downarrow$} \\
\midrule
1 & $\{8\}$               & 0.381 & 0.412 & 0.158 & 0.248 \\
\midrule
2 & $\{4,8\}$             & 0.379 & 0.410 & 0.157 & 0.247 \\
  & $\{2,8\}$             & 0.379 & 0.411 & 0.157 & 0.247 \\
  & $\{1,8\}$             & 0.380 & 0.409 & 0.157 & 0.247 \\
\midrule
3 & $\{2,4,8\}$           & 0.380 & 0.411 & 0.158 & 0.248 \\
  & $\{1,4,8\}$           & 0.378 & 0.409 & 0.158 & 0.247 \\
  & $\{3,6,8\}$           & 0.379 & 0.410 & 0.156 & 0.248 \\
\midrule
4 & $\{1,2,4,8\}$         & \textbf{0.376} & \textbf{0.408} & \textbf{0.154} & \textbf{0.245} \\
  & $\{1,3,6,8\}$         & 0.379 & 0.409 & 0.159 & 0.247 \\
\midrule
5 & $\{1,2,4,6,8\}$       & 0.379 & 0.410 & 0.159 & 0.248 \\
\bottomrule
\end{tabular}
\end{adjustbox}
\caption{\textbf{Scale token schedule ablation.} The 4-step schedule $\{1,2,4,8\}$ performs best and is used as default.}
\label{tab:ab_scale_list}
\end{table}

\noindent\textbf{Effect of the Refine Loss}
\label{sec:ablation_refine_loss}
The refine loss $\mathcal{L}_{\text{refine}}$ propagates gradients into the shared encoder and decoder, aligning token representations at each scale with their downscaled targets. As Table~\ref{tab:ablation_refine_loss} shows, removing it degrades accuracy across all datasets. Full LLM backbone fine-tuning yields only marginal gains at $15.21\times$ the training cost, confirming that scale-wise supervision via $\mathcal{L}_{\text{refine}}$ is both effective and efficient.

\begin{table}[hpt]
\centering
\small
\setlength{\tabcolsep}{5pt}
\renewcommand{\arraystretch}{1.05}
\begin{adjustbox}{max width=0.9\linewidth}
\begin{tabular}{lccccc}
\toprule
\textbf{Method} & \textbf{ETTh1} & \textbf{ECL} & \textbf{Weather} & \textbf{Traffic} & \textbf{Train Time} \\
\midrule
SCALER (default)            & \textbf{0.376} & \textbf{0.154} & \textbf{0.217} & \textbf{0.367} & $1.00\times$ \\
w/o refine loss             & 0.380          & 0.158          & 0.220          & 0.371          & $1.00\times$ \\
w/ LLM backbone fine-tuning & 0.376          & 0.152          & 0.218          & 0.368          & $15.21\times$ \\
\bottomrule
\end{tabular}
\end{adjustbox}
\caption{\textbf{Effect of refine loss and LLM fine-tuning (MSE$\downarrow$).}}
\label{tab:ablation_refine_loss}
\end{table}

\begin{table*}[hpbt]

\centering
\caption{\textbf{Full long-term forecasting results.} }
\label{tab:ltsf_full}
\setlength{\tabcolsep}{2.0pt}
\renewcommand{\arraystretch}{1.02}
\begin{adjustbox}{width=\textwidth}
\begin{tabular}{@{}l|l|cc|cc|cc|cc|cc|cc|cc|cc|cc|cc|cc|cc@{}}
\toprule
                         &                  & \multicolumn{4}{c}{Test-Time Scaling}                                                     & \multicolumn{12}{|c}{LLM-based Model}                                                                                                                                                                                         & \multicolumn{8}{|c}{Other Deep Learning   Model}                                                                                                           \\ \midrule
Dataset                  & Length & \multicolumn{2}{c}{SCALER (Ours)}           & \multicolumn{2}{c}{TimeReasoner}            & \multicolumn{2}{c}{LVICL}           & \multicolumn{2}{c}{AutoTimes}       & \multicolumn{2}{c}{TimeLLM}       & \multicolumn{2}{c}{FPT}             & \multicolumn{2}{c}{Chronos}           & \multicolumn{2}{c}{TimeFM}          & \multicolumn{2}{c}{iTransformer}    & \multicolumn{2}{c}{DLinear}         & \multicolumn{2}{c}{PatchTST}            & \multicolumn{2}{c}{TimesNet}        \\
                         &                  & MSE                  & MAE                  & MSE                  & MAE                  & MSE              & MAE              & MSE              & MAE              & MSE              & MAE            & MSE              & MAE              & MSE            & MAE                  & MSE              & MAE              & MSE              & MAE              & MSE              & MAE              & MSE              & MAE                  & MSE              & MAE              \\ \midrule
\multirow{5}{*}{ETTh1}   & 96               & \textbf{0.345}       & \textbf{0.386}       & 0.36                 & 0.39                 & {\ul 0.351}      & {\ul 0.389}      & 0.36             & 0.4              & 0.362            & 0.392          & 0.376            & 0.397            & 0.369          & 0.409                & 0.384            & 0.426            & 0.386            & 0.405            & 0.375            & 0.399            & 0.37             & 0.399                & 0.384            & 0.402            \\
                         & 192              & \textbf{0.374}       & \textbf{0.406}       & 0.387                & 0.412                & {\ul 0.379}      & {\ul 0.408}      & 0.388            & 0.419            & 0.398            & 0.418          & 0.416            & 0.418            & 0.399          & 0.429                & 0.415            & 0.447            & 0.422            & 0.439            & 0.405            & 0.416            & 0.413            & 0.421                & 0.557            & 0.436            \\
                         & 336              & \textbf{0.385}       & \textbf{0.413}       & 0.425                & {\ul 0.417}          & {\ul 0.392}      & {\ul 0.417}      & 0.401            & 0.429            & 0.43             & 0.427          & 0.442            & 0.433            & 0.413          & 0.439                & 0.429            & 0.456            & 0.444            & 0.457            & 0.439            & 0.443            & 0.422            & 0.436                & 0.491            & 0.469            \\
                         & 720              & \textbf{0.399}       & \textbf{0.425}       & 0.439                & 0.449                & {\ul 0.402}      & {\ul 0.434}      & 0.406            & 0.44             & 0.442            & 0.457          & 0.477            & 0.456            & 0.423          & 0.457                & 0.44             & 0.475            & 0.5              & 0.498            & 0.472            & 0.49             & 0.447            & 0.466                & 0.521            & 0.5              \\
                         & Avg              & \textbf{0.376}       & \textbf{0.408}       & 0.403                & 0.417                & {\ul 0.381}      & {\ul 0.412}      & 0.389            & 0.422            & 0.408            & 0.423          & 0.427            & 0.426            & 0.401          & 0.434                & 0.417            & 0.451            & 0.438            & 0.45             & 0.423            & 0.437            & 0.413            & 0.431                & 0.458            & 0.45             \\ \midrule
\multirow{5}{*}{ETTh2}   & 96               & \textbf{0.265}       & \textbf{0.33}        & 0.283                & 0.336                & {\ul 0.27}       & {\ul 0.332}      & 0.282            & 0.342            & 0.288            & 0.341          & 0.287            & 0.341            & 0.271          & 0.333                & 0.277            & 0.34             & 0.304            & 0.36             & 0.289            & 0.353            & 0.274            & 0.336                & 0.34             & 0.374            \\
                         & 192              & {\ul 0.327}          & {\ul 0.376}          & 0.346                & 0.388                & 0.333            & 0.378            & 0.348            & 0.387            & 0.351            & 0.389          & 0.35             & 0.383            & \textbf{0.326} & \textbf{0.371}       & 0.334            & 0.38             & 0.377            & 0.403            & 0.383            & 0.418            & 0.339            & 0.379                & 0.402            & 0.414            \\
                         & 336              & \textbf{0.322}       & \textbf{0.371}       & 0.352                & 0.398                & {\ul 0.325}      & {\ul 0.377}      & 0.365            & 0.412            & 0.362            & 0.401          & 0.373            & 0.395            & 0.338          & 0.394                & 0.34             & 0.393            & 0.405            & 0.429            & 0.448            & 0.465            & 0.329            & 0.38                 & 0.452            & 0.452            \\
                         & 720              & \textbf{0.368}       & {\ul 0.415}          & 0.411                & 0.432                & 0.375            & 0.416            & 0.412            & 0.44             & 0.415            & 0.44           & 0.401            & 0.443            & {\ul 0.372}    & \textbf{0.409}       & 0.384            & 0.426            & 0.443            & 0.464            & 0.605            & 0.551            & 0.379            & 0.422                & 0.462            & 0.468            \\
                         & Avg              & \textbf{0.321}       & \textbf{0.373}       & 0.348                & 0.389                & {\ul 0.326}      & {\ul 0.376}      & 0.352            & 0.395            & 0.354            & 0.393          & 0.353            & 0.391            & 0.327          & 0.377                & 0.334            & 0.385            & 0.382            & 0.414            & 0.431            & 0.446            & 0.33             & 0.379                & 0.414            & 0.427            \\ \midrule
\multirow{5}{*}{ETTm1}   & 96               & {\ul 0.265}          & 0.342                & 0.281                & {\ul \textbf{0.337}} & 0.269            & 0.343            & 0.274            & 0.343            & 0.284            & 0.341          & 0.301            & 0.343            & \textbf{0.264} & {\ul \textbf{0.337}} & 0.28             & 0.357            & 0.312            & 0.366            & 0.299            & 0.343            & 0.29             & 0.342                & 0.338            & 0.375            \\
                         & 192              & \textbf{0.302}       & 0.368                & 0.32                 & 0.368                & 0.309            & 0.371            & 0.316            & 0.37             & 0.327            & {\ul 0.363}    & 0.348            & \textbf{0.362}   & {\ul 0.303}    & 0.364                & 0.321            & 0.386            & 0.347            & 0.385            & 0.335            & 0.365            & 0.332            & 0.369                & 0.374            & 0.387            \\
                         & 336              & {\ul 0.341}          & 0.386                & 0.363                & \textbf{0.379}       & 0.343            & 0.39             & 0.344            & 0.39             & 0.368            & 0.387          & 0.386            & 0.397            & \textbf{0.337} & {\ul 0.383}          & 0.357            & 0.405            & 0.379            & 0.404            & 0.369            & 0.386            & 0.366            & 0.392                & 0.41             & 0.411            \\
                         & 720              & {\ul 0.388}          & \textbf{0.398}       & 0.413                & 0.416                & 0.391            & 0.408            & 0.392            & 0.418            & 0.423            & 0.419          & 0.43             & 0.427            & \textbf{0.384} & {\ul 0.401}          & 0.406            & 0.424            & 0.441            & 0.442            & 0.425            & 0.421            & 0.416            & 0.42                 & 0.478            & 0.45             \\
                         & Avg              & {\ul 0.324}          & 0.374                & 0.344                & {\ul 0.373}          & 0.328            & 0.378            & 0.332            & 0.38             & 0.35             & 0.378          & 0.366            & 0.382            & \textbf{0.322} & \textbf{0.371}       & 0.341            & 0.393            & 0.37             & 0.399            & 0.357            & 0.378            & 0.351            & 0.38                 & 0.4              & 0.406            \\ \midrule
\multirow{5}{*}{ETTm2}   & 96               & \textbf{0.148}       & \textbf{0.245}       & 0.157                & 0.248                & {\ul 0.155}      & {\ul 0.246}      & 0.159            & 0.25             & 0.168            & 0.25           & 0.172            & 0.258            & 0.159          & 0.252                & 0.157            & 0.249            & 0.179            & 0.271            & 0.167            & 0.269            & 0.165            & 0.255                & 0.187            & 0.267            \\
                         & 192              & \textbf{0.204}       & \textbf{0.281}       & {\ul 0.208}          & 0.291                & 0.21             & {\ul 0.282}      & 0.213            & 0.285            & 0.216            & 0.294          & 0.231            & 0.287            & 0.215          & 0.288                & 0.212            & 0.285            & 0.242            & 0.313            & 0.224            & 0.303            & 0.22             & 0.292                & 0.249            & 0.309            \\
                         & 336              & \textbf{0.253}       & \textbf{0.311}       & {\ul 0.257}          & 0.319                & 0.258            & {\ul 0.317}      & 0.261            & 0.323            & 0.269            & 0.323          & 0.282            & 0.329            & 0.264          & 0.324                & 0.261            & 0.32             & 0.288            & 0.344            & 0.281            & 0.342            & 0.274            & 0.329                & 0.321            & 0.351            \\
                         & 720              & \textbf{0.332}       & \textbf{0.372}       & 0.357                & 0.384                & {\ul 0.335}      & {\ul 0.378}      & 0.341            & 0.381            & 0.363            & 0.387          & 0.373            & 0.387            & 0.343          & 0.387                & 0.338            & 0.382            & 0.378            & 0.397            & 0.397            & 0.421            & 0.362            & 0.385                & 0.408            & 0.403            \\
                         & Avg              & \textbf{0.234}       & \textbf{0.302}       & 0.245                & 0.311                & {\ul 0.239}      & {\ul 0.306}      & 0.243            & 0.31             & 0.254            & 0.314          & 0.265            & 0.315            & 0.245          & 0.313                & 0.242            & 0.309            & 0.272            & 0.331            & 0.267            & 0.333            & 0.255            & 0.315                & 0.291            & 0.333            \\ \midrule
\multirow{5}{*}{ECL}     & 96               & \textbf{0.12}        & {\ul \textbf{0.222}} & 0.128                & {\ul \textbf{0.222}} & {\ul 0.126}      & 0.224            & 0.129            & 0.225            & 0.131            & 0.224          & 0.139            & 0.238            & 0.131          & 0.233                & 0.129            & 0.23             & 0.132            & 0.227            & 0.153            & 0.237            & 0.129            & {\ul \textbf{0.222}} & 0.168            & 0.272            \\
                         & 192              & \textbf{0.143}       & \textbf{0.231}       & {\ul 0.146}          & 0.24                 & 0.148            & {\ul 0.235}      & 0.147            & 0.241            & 0.152            & 0.241          & 0.153            & 0.251            & 0.154          & 0.244                & 0.152            & 0.241            & 0.153            & 0.249            & 0.152            & 0.249            & 0.147            & 0.24                 & 0.184            & 0.289            \\
                         & 336              & {\ul 0.158}          & {\ul 0.246}          & \textbf{0.153}       & \textbf{0.242}       & 0.161            & 0.252            & 0.162            & 0.258            & 0.16             & 0.248          & 0.169            & 0.266            & 0.167          & 0.262                & 0.165            & 0.259            & 0.167            & 0.262            & 0.169            & 0.267            & 0.163            & 0.259                & 0.198            & 0.3              \\
                         & 720              & 0.194                & \textbf{0.281}       & \textbf{0.189}       & 0.294                & 0.196            & {\ul 0.282}      & 0.199            & 0.288            & {\ul 0.192}      & 0.298          & 0.206            & 0.297            & 0.204          & 0.293                & 0.201            & 0.29             & 0.254            & 0.335            & 0.233            & 0.344            & 0.197            & 0.29                 & 0.22             & 0.32             \\
                         & Avg              & {\ul \textbf{0.154}} & \textbf{0.245}       & {\ul \textbf{0.154}} & 0.25                 & 0.158            & {\ul 0.248}      & 0.159            & 0.253            & 0.159            & 0.253          & 0.167            & 0.263            & 0.164          & 0.258                & 0.162            & 0.255            & 0.161            & 0.256            & 0.177            & 0.274            & 0.159            & 0.253                & 0.192            & 0.295            \\ \midrule
\multirow{5}{*}{Weather} & 96               & {\ul 0.141}          & \textbf{0.193}       & \textbf{0.136}       & 0.198                & 0.146            & {\ul 0.196}      & 0.153            & 0.203            & 0.147            & 0.201          & 0.162            & 0.212            & 0.145          & 0.198                & 0.151            & 0.202            & 0.163            & 0.211            & 0.152            & 0.237            & 0.149            & 0.198                & 0.172            & 0.22             \\
                         & 192              & {\ul 0.186}          & 0.238                & \textbf{0.181}       & \textbf{0.228}       & 0.19             & 0.24             & 0.201            & 0.25             & 0.189            & {\ul 0.234}    & 0.204            & 0.248            & 0.188          & 0.242                & 0.196            & 0.248            & 0.205            & 0.25             & 0.22             & 0.282            & 0.194            & 0.241                & 0.219            & 0.261            \\
                         & 336              & \textbf{0.234}       & {\ul \textbf{0.277}} & 0.258                & {\ul \textbf{0.277}} & 0.24             & 0.28             & 0.256            & 0.293            & 0.262            & 0.279          & 0.254            & 0.286            & {\ul 0.238}    & 0.281                & 0.248            & 0.289            & 0.254            & 0.289            & 0.265            & 0.319            & 0.245            & 0.282                & 0.28             & 0.306            \\
                         & 720              & 0.306                & 0.32                 & \textbf{0.296}       & \textbf{0.312}       & 0.312            & 0.33             & 0.331            & 0.345            & {\ul 0.304}      & {\ul 0.316}    & 0.326            & 0.337            & 0.309          & 0.331                & 0.322            & 0.34             & 0.329            & 0.34             & 0.323            & 0.362            & 0.314            & 0.334                & 0.365            & 0.359            \\
                         & Avg              & \textbf{0.217}       & {\ul 0.257}          & {\ul 0.218}          & \textbf{0.254}       & 0.224            & 0.262            & 0.235            & 0.273            & 0.225            & {\ul 0.257}    & 0.237            & 0.27             & 0.22           & 0.263                & 0.229            & 0.27             & 0.238            & 0.272            & 0.24             & 0.3              & 0.226            & 0.264                & 0.259            & 0.287            \\ \midrule
\multirow{5}{*}{Traffic} & 96               & \textbf{0.334}       & \textbf{0.244}       & 0.357                & 0.247                & {\ul 0.338}      & {\ul 0.245}      & 0.343            & 0.248            & 0.362            & 0.248          & 0.388            & 0.282            & 0.343          & 0.249                & 0.35             & 0.253            & 0.351            & 0.257            & 0.41             & 0.282            & 0.36             & 0.249                & 0.593            & 0.321            \\
                         & 192              & \textbf{0.358}       & 0.25                 & 0.37                 & {\ul 0.248}          & {\ul 0.36}       & 0.251            & 0.362            & 0.257            & 0.374            & \textbf{0.247} & 0.407            & 0.29             & 0.366          & 0.255                & 0.372            & 0.26             & 0.364            & 0.265            & 0.423            & 0.287            & 0.379            & 0.256                & 0.617            & 0.333            \\
                         & 336              & \textbf{0.378}       & \textbf{0.254}       & 0.381                & 0.264                & 0.382            & {\ul 0.259}      & {\ul 0.379}      & 0.266            & 0.385            & 0.271          & 0.412            & 0.294            & 0.388          & 0.263                & 0.395            & 0.268            & 0.382            & 0.273            & 0.436            & 0.296            & 0.392            & 0.264                & 0.629            & 0.333            \\
                         & 720              & \textbf{0.399}       & \textbf{0.276}       & 0.421                & {\ul 0.281}          & {\ul 0.401}      & 0.285            & 0.413            & 0.284            & 0.43             & 0.288          & 0.45             & 0.312            & 0.407          & 0.289                & 0.415            & 0.295            & 0.42             & 0.292            & 0.466            & 0.315            & 0.432            & 0.286                & 0.64             & 0.33             \\
                         & Avg              & \textbf{0.367}       & \textbf{0.256}       & 0.382                & {\ul 0.259}          & {\ul 0.37}       & 0.26             & 0.374            & 0.264            & 0.388            & 0.264          & 0.414            & 0.294            & 0.376          & 0.264                & 0.383            & 0.269            & 0.379            & 0.272            & 0.434            & 0.295            & 0.391            & 0.264                & 0.62             & 0.336            \\ \midrule
                         & 1st Count        & \textbf{19 }                  & \textbf{19}                   & {\ul 5}                    & {\ul 7}                    & 0 & 0 & 0 & 0 & 0 & 1              & 0 & 0 & 4              & 4                    & 0 & 0 & 0 & 0 & 0 & 0 & 0 & 1                    & 0 & 0 \\ \bottomrule
\end{tabular}
\end{adjustbox}

\end{table*}

\noindent\textbf{LLM Refiner vs.\ Smaller Model Refiners}
\label{sec:llm_vs_small}
We replace the LLM refiner with PatchTST~\cite{patchtst} and iTransformer~\cite{itransformer}, keeping all else unchanged. Both alternatives outperform their standalone counterparts, confirming the value of the coarse-to-fine structure. However, LLaMA-7B consistently achieves the best accuracy (Table~\ref{tab:llm_vs_small}), indicating that the LLM's pretrained generative prior provides additional coherence that smaller models cannot match.

\begin{table}[h]
\centering
\small
\setlength{\tabcolsep}{7pt}
\renewcommand{\arraystretch}{1.05}
\begin{adjustbox}{max width=0.9\linewidth}
\begin{tabular}{lcccc}
\toprule
\textbf{Stage-II Refiner}   & \multicolumn{2}{c}{\textbf{ETTh1}} & \multicolumn{2}{c}{\textbf{ECL}} \\
                             & MSE$\downarrow$ & MAE$\downarrow$  & MSE$\downarrow$ & MAE$\downarrow$ \\
\midrule
SCALER (LLaMA-7B, default)  & \textbf{0.376} & \textbf{0.408}   & \textbf{0.154} & \textbf{0.245}  \\
SCALER (PatchTST)           & 0.401          & 0.425            & 0.159          & 0.254           \\
SCALER (iTransformer)       & 0.424          & 0.443            & 0.158          & 0.251           \\
\bottomrule
\end{tabular}
\end{adjustbox}
\caption{\textbf{LLM vs.\ smaller refiners.} LLaMA-7B outperforms both alternatives, showing the pretrained prior contributes beyond the coarse-to-fine structure alone.}
\label{tab:llm_vs_small}
\end{table}

\section{Further Analysis}

\noindent\textbf{Intermediate-Stage Analysis}
\label{sec:intermediate_stages}
Table~\ref{tab:intermediate_stages} shows that MSE decreases monotonically from Stage~1 (coarse shape) through Stage~4 (full refinement) on both ETTh1 and ECL, confirming that each refinement step contributes meaningfully rather than redundantly.

\begin{table}[h]
\centering
\small
\setlength{\tabcolsep}{8pt}
\renewcommand{\arraystretch}{1.05}
\begin{adjustbox}{max width=0.9\linewidth}
\begin{tabular}{lcc}
\toprule
\textbf{Stage}                 & \textbf{ETTh1 (MSE$\downarrow$)} & \textbf{ECL (MSE$\downarrow$)} \\
\midrule
Stage-1 (coarse shape only)    & 0.391 & 0.168 \\
Stage-2                        & 0.387 & 0.165 \\
Stage-3                        & 0.384 & 0.162 \\
SCALER (Stage-4, default)      & \textbf{0.376} & \textbf{0.154} \\
\bottomrule
\end{tabular}
\end{adjustbox}
\caption{\textbf{Intermediate-stage MSE.} Performance improves monotonically, validating each refinement step.}
\label{tab:intermediate_stages}
\end{table}

\noindent\textbf{Discussion: downscaling and aliasing.}
The normalise-then-interpolate operator risks aliasing if the downsampling grid coincides with zeros of a periodic component. The default $8\times$ ratio ($T{=}96{\to}12$) retains sufficient structure for our benchmarks, but for high-frequency signals a smaller ratio ($4\times$ or $6\times$) is advisable. Adaptive ratio selection and explicit trend/seasonality encoding are left for future work.

\section{Full Long-Term Forecasting Results}
\label{sec:full_results}

Table~\ref{tab:ltsf_full} reports full results on seven benchmarks across horizons $\{96,192,336,720\}$. SCALER achieves consistently strong performance, particularly at longer horizons, winning 19 of 28 dataset--horizon pairs on both MSE and MAE. Baseline results are taken from \cite{llm-lvicl}; TimeReasoner is rerun under our setup.

\end{document}